\documentclass{article} % For LaTeX2e
\usepackage{iclr2027_conference,times}
\let\cite\citep

\usepackage{amsmath,amsfonts,bm}

\def\eqref#1{equation~\ref{#1}}
\def\1{\bm{1}}

\DeclareMathAlphabet{\mathsfit}{\encodingdefault}{\sfdefault}{m}{sl}
\SetMathAlphabet{\mathsfit}{bold}{\encodingdefault}{\sfdefault}{bx}{n}

\usepackage{tikz}
\usetikzlibrary{positioning,arrows.meta,calc,fit,backgrounds}

\usepackage{hyperref}
\usepackage{url}

\usepackage[framemethod=TikZ]{mdframed}
\newmdenv[
leftline=true, topline=false, bottomline=false, rightline=false,
linewidth=2pt, linecolor=black!25,
innerleftmargin=10pt, innerrightmargin=0pt,
innertopmargin=4pt, innerbottommargin=4pt,
skipabove=\medskipamount, skipbelow=\medskipamount
]{barquote}

\newif\ifanonym
\anonymfalse 
\usepackage{graphicx}
\usepackage{cleveref}
\usepackage{amsthm}
\usepackage{thmtools}
\usepackage{multirow}
\usepackage{booktabs} 
\usepackage{diagbox}
\usepackage{enumerate}
\usepackage{csquotes}
\usepackage[table]{xcolor}   
\usepackage{subcaption}

\usepackage{xcolor}
\definecolor{quotemark}{gray}{0.65}

\newcommand{\bigquote}[1]{%
	{\leftskip=2em plus 1fil \rightskip=2em plus 1fil
		\parfillskip=0pt \parindent=0pt
		\leavevmode
		\llap{\smash{\color{quotemark}%
				\raisebox{-1em}{\scalebox{2.4}{\textquotedblleft}}\hspace{0.1em}}}%
		#1%
		\rlap{\smash{\color{quotemark}%
				\hspace{0.1em}\raisebox{-1em}{\scalebox{2.4}{\textquotedblright}}}}%
		\par}%
}

\definecolor{special}{HTML}{6B4FA0}    % muted violet, text
\definecolor{specialbg}{HTML}{EFEAF6}  % same hue, ~8% tint

\newcommand{\spec}[1]{\textcolor{special}{#1}\textsuperscript{\dag}}
\newcommand{\env}[1]{\texttt{#1}}

\usepackage{mathtools}
\usetikzlibrary{arrows.meta,positioning,fit,backgrounds,calc,shapes.geometric}

\makeatletter
\renewcommand\paragraph{\@startsection{paragraph}{4}{\z@}%
	{0.5ex \@plus 0.2ex \@minus 0.2ex}% space above (ICLR default is ~1ex plus 0.5ex)
	{-0.8em}%                          
	{\normalsize\bfseries}}
\makeatother

\definecolor{ink}{HTML}{1F2933}
\definecolor{machinefill}{HTML}{E6EAF0}
\definecolor{agentfill}{HTML}{CFE3F5}
\definecolor{storefill}{HTML}{F6E7C8}
\definecolor{accent}{HTML}{2B5D8A}

\tikzset{
	>=Stealth,
	every node/.style={font=\sffamily,text=ink},
	title/.style={font=\sffamily\Large\bfseries,text=ink},
	lbl/.style={font=\sffamily\small,text=ink},
	flow/.style={->,thick,draw=ink},
	box/.style={draw=ink,thick,rounded corners=2pt,fill=white},
	group/.style={draw=accent,thick,rounded corners=8pt,fill=accent!6,inner sep=16pt},
	pool/.style={draw=ink!55,thick,dashed,rounded corners=8pt,inner sep=14pt},
	machine/.pic={
		\draw[thick,fill=machinefill,rounded corners=1.5pt] (-0.75,-0.55) rectangle (0.75,0.45);
		\draw[thick,fill=machinefill,rounded corners=1.5pt] (-0.55,0.45) rectangle (0.05,0.85);
		\draw[thick,fill=white] (-0.46,0.56) rectangle (-0.04,0.77);
		\draw[thick] (0.40,0.45) -- (0.40,1.00) -- (0.66,1.00);
		\foreach \a in {0,45,...,315}{%
			\draw[thick] ([shift=(\a:0.20)]0.34,-0.05) -- ([shift=(\a:0.30)]0.34,-0.05);}
		\draw[thick,fill=white] (0.34,-0.05) circle (0.20);
		\draw[thick,fill=machinefill] (0.34,-0.05) circle (0.06);
		\draw[thick,fill=white] (-0.50,-0.22) circle (0.14);
		\draw[thick,fill=white] (-0.12,-0.22) circle (0.14);
		\draw[thick] (-0.50,-0.08) -- (-0.12,-0.08);
		\draw[thick] (-0.50,-0.36) -- (-0.12,-0.36);
		\draw[thick] (-0.50,-0.55) -- (-0.50,-0.78);  \draw[thick] (-0.70,-0.78) -- (-0.30,-0.78);
		\draw[thick] ( 0.50,-0.55) -- ( 0.50,-0.78);  \draw[thick] ( 0.30,-0.78) -- ( 0.70,-0.78);
		\node[inner sep=0,minimum width=1.60cm,minimum height=1.85cm] (-c) at (0,0.05) {};
	},
	robot/.pic={
		\draw[thick] (0,0.95) -- (0,1.16);  \draw[thick,fill=ink] (0,1.23) circle (0.07);
		\draw[thick,fill=agentfill,rounded corners=4pt] (-0.45,0.42) rectangle (0.45,0.95);
		\draw[thick,fill=white] (-0.19,0.72) circle (0.10);
		\draw[thick,fill=white] ( 0.19,0.72) circle (0.10);
		\fill[ink] (-0.19,0.72) circle (0.05);   \fill[ink] (0.19,0.72) circle (0.05);
		\draw[thick] (-0.16,0.55) -- (0.16,0.55);
		\draw[thick] (0,0.42) -- (0,0.34);
		\draw[thick,fill=agentfill,rounded corners=4pt] (-0.52,-0.52) rectangle (0.52,0.34);
		\draw[thick,fill=white,rounded corners=1pt] (-0.24,-0.28) rectangle (0.24,-0.02);
		\draw[thick] (-0.52,0.14) -- (-0.82,-0.08) -- (-0.82,-0.34);
		\draw[thick] ( 0.52,0.14) -- ( 0.82,-0.08) -- ( 0.82,-0.34);
		\draw[thick] (-0.26,-0.52) -- (-0.26,-0.76); \draw[thick] (-0.44,-0.76) -- (-0.08,-0.76);
		\draw[thick] ( 0.26,-0.52) -- ( 0.26,-0.76); \draw[thick] ( 0.08,-0.76) -- ( 0.44,-0.76);
		\node[inner sep=0,minimum width=1.70cm,minimum height=2.10cm] (-c) at (0,0.20) {};
	}
}

\declaretheorem[style=assumptionstyle, name=Assumption]{assumption}
\declaretheorem[style=theoremstyle, name=Theorem]{theorem}
\declaretheorem[style=theoremstyle, name=Corollary]{corollary}

\crefname{assumption}{assumption}{assumptions}
\Crefname{assumption}{Assumption}{Assumptions} 

\title{Space-sampled Value Decay:\\ Forgetting Mechanisms for\\ Non-stationary Reinforcement Learning}

\author{Felix Störck$^{1}$, Philipp Hartmann$^{1}$, Fabian Hinder$^{1}$, Klaus Neumann$^{1,2}$, Barbara Hammer$^{1}$ \medskip \\  $^1$ Bielefeld University, CITEC \quad $^2$ Fraunhofer IOSB-INA, Lemgo }

\ifanonym

\else
	\iclrfinalcopy
\fi
\begin{document}
	
	\maketitle
	\ifanonym
	
	\else
		\lhead{Preprint. An earlier version (v1) was presented at EIML@ICML2026 (non-archival).} 
	\fi
	\begin{abstract} 
		Reinforcement Learning agents deployed on physical systems must adapt continually, since degradation and shifting environment conditions change the dynamics (they \emph{drift}) over time. 
		In the hardest version of this problem, the agent interacts with a single system that might drift at every timestep, leaving no opportunity to revisit past conditions -- a setting we call Single Environment, One-Shot Non-Stationary Reinforcement Learning (SEOS-NSRL).
		We argue that this setting calls for selective forgetting rather than re-learning, and introduce Space-sampled Value Decay (SsVD), which pulls value estimates of randomly chosen elements of the state space to a baseline value, so that outdated information in non visited regions is discarded.
		SsVD does not require resetting or change-point detection and plugs into modern off-policy algorithms; we integrate it into Soft Actor Critic and Deep Q-Networks.
		Across 6 non-stationary environments, SsVD improves upon its direct base algorithms and attains the best mean rank across all. The SsVD mechanism can also induce optimism which we show on hard-exploration tasks, although we investigate the connection here only briefly.
	\end{abstract}
	
	\section{Introduction}
\begin{figure}[t]
		\centering
		\includegraphics[width=1\textwidth]{
			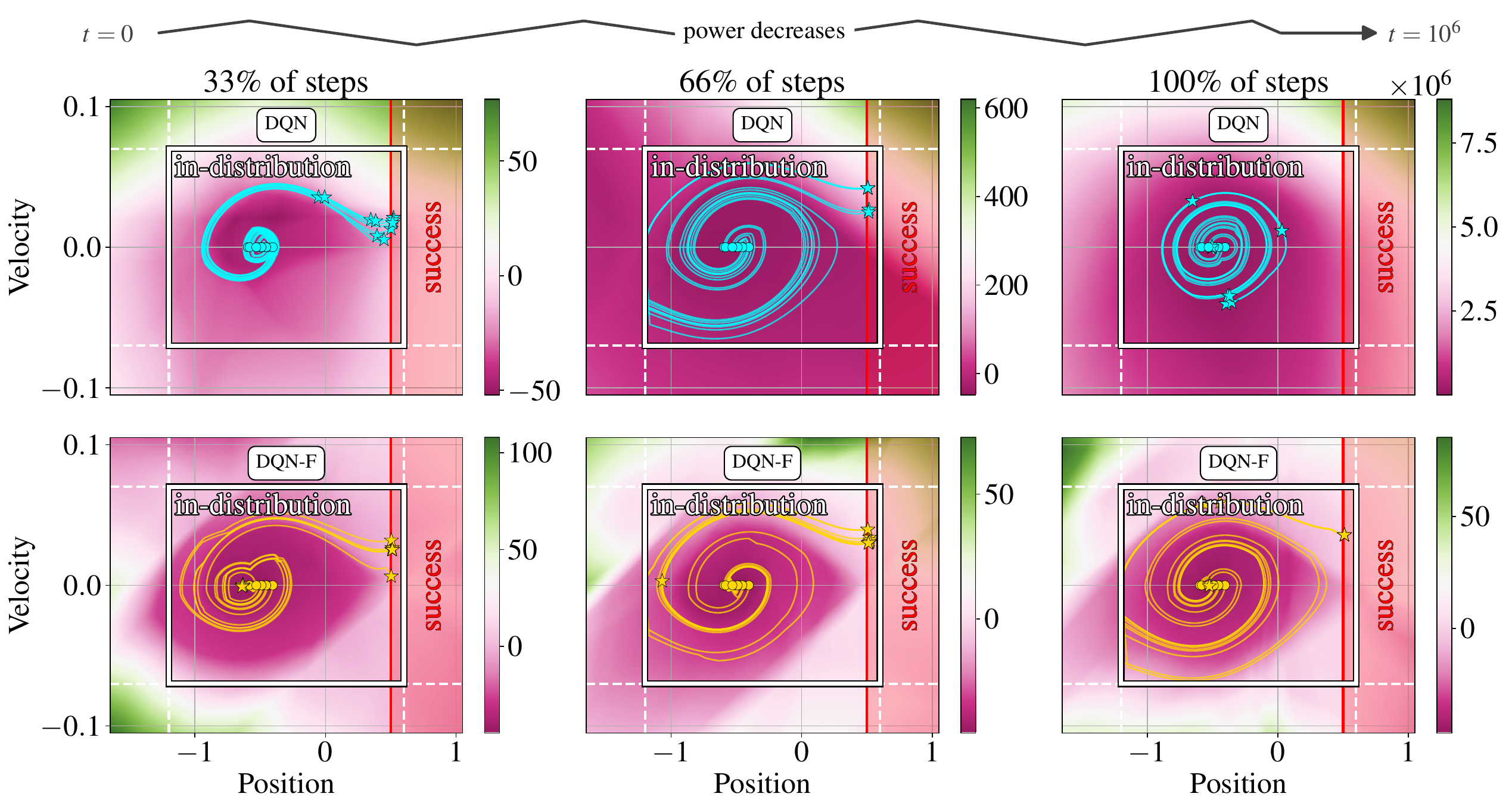}
		\caption{Heatmap snapshots of the learned action-value function ($\max_{a}Q(s,a)$ over the state-space) by DQN (top, weight decay enabled) and DQN-F (DQN + SsVD, bottom, weight decay disabled, baseline $b=0$), on a non-stationary MountainCar environment where the engine power decreases across time. Sample trajectories of different rollouts at the same intermediate steps (colored lines), start position (circle marker) and end position (star marker). Total of 1M training steps. A rollout is successful if a position with $x\geq0.5$ is reached (red shaded region, right side).}
		\label{fig:visabs}
	\end{figure}

	Humans are inherently capable of adapting to changing environmental conditions.
	Replicating this behavior in intelligent agents is however challenging even though an ever increasing focus is placed upon the need for such life-long learning capabilities of agents based on experience \cite{silver_welcome_2025}.
	As of today, most modern Reinforcement Learning architectures lack mechanisms for adapting to changes (we refer to such changes as \emph{drift}) in the environment -- see \Cref{fig:visabs}.
	We coarsely refer to this problem setting as \emph{Non-Stationary Reinforcement Learning} (NSRL). 
	In particular, we are interested in environmental drift (e.g. of the dynamics) that necessitates a policy adaptation.
	
	Whilst the setting of non-stationary Reinforcement Learning has been acknowledged for a while \cite{sutton_dyna_1991, sutton_role_2007}, the limited view of clearly distinct, finite set of \emph{tasks} is very prevalent \citep[see][for an overview]{khetarpal_towards_2022-1}.
	Recently, arguments for a more general view of NSRL with continuous or ever-going changes have been voiced \cite{silver_welcome_2025,behdin_position_2026}.
	We take and further refine this setting as \emph{single environment, one-shot} NSRL assuming that the agent can only interact with a single environment which in the worst case changes at every timestep -- thus being \emph{one-shot}.
	This corresponds to the setting when RL agents are a) deployed \cite{behdin_position_2026} or b) interact with real-world systems \cite{dulac-arnold_challenges_2021}.
	
	The concept of forgetting mechanisms plays a vital role in the design of machine learning algorithms (such as LSTMs \cite{hochreiter_long_1997}) but is also a key aspect of neurology research \cite{richards_persistence_2017,ryan_forgetting_2022}:
	\begin{barquote}
		\begin{center}
			\bigquote{In other words, degrading stored information contained in a memory allows organisms to behave more flexibly and promotes better memory-guided decision-making.}
		\end{center}
		\raggedleft --- \cite[p.~179]{ryan_forgetting_2022}
	\end{barquote}
	This work takes inspiration from ``Q-Learning with forgetting'' \cite{ito_validation_2009,beron_mice_2022} designed to model the behaviour of rodents in non-stationary conditions and extend the key concepts to modern Reinforcement Learning architectures.
	Our contributions are the following:
	\begin{itemize}\setlength\itemsep{0.25em} 
		\item We introduce \emph{Space-sampled Value Decay} (SsVD), a bio-inspired forgetting mechanism that samples states from the state space pulling their values toward a baseline $b$. It adds one auxiliary loss, needs no resets or drift detection, and plugs into any off-policy RL agent.
		
		\item 	Across 6 non-stationary environments, SsVD improves upon its direct base algorithms and attains the best mean rank across all tested methods. We  provide qualitative insights about SsVD by performing a visual analysis on \env{MountainCar}.
		An optimistic baseline $b$ turns the same mechanism into an exploration bonus which we demonstrate on 3 environments.
	\end{itemize}

	\section{Problem Setting} \label{sec:problem_setting}
	
	\paragraph*{Reinforcement Learning}
	Reinforcement Learning assumes an underlying Markov Decision Process (MDP)
	which is defined by a tuple $(\mathcal{S}, \mathcal{A}, P, R, \gamma)$, where $\mathcal{S}$ is the state space, $\mathcal{A}$ is the action space, $P(s'|s,a)$ is the transition dynamics, $R(s,a)$ is the reward function, and $\gamma \in [0,1)$ is a discount factor. At each timestep $t$, an agent observes state $s_t \in \mathcal{S}$, selects action $a_t \in \mathcal{A}$ according to a policy $\pi(a|s)$, and receives reward $r_t = R(s_t, a_t)$. The goal is to find a policy $\pi^*$ that maximizes the expected discounted return $\mathbb{E}_\pi\left[\sum_{t=0}^{\infty} \gamma^t r_t\right]$.
	
	\paragraph*{Q-Learning}\cite{watkins_technical_1992}
	Default Q-Learning assumes a discrete state and action space with the goal of learning an action-value function that provides the expected discounted return of taking action $a_t$ in state $s_t$ from timestep $t$ onwards.
	The idea is to update the current action-value estimate based on the immediately received reward $r_t$ and a discounted expected future return provided by bootstrapping the action-value function (denoted by $Q$) as follows:
	\begin{align}\label{eq:QL}
		Q(s_t, a_t) \gets  (1-\alpha)&Q(s_t, a_t) + \alpha [r_{t} + \gamma \max_{\tilde{a}} Q(s_{t+1},\tilde{a})]. 
	\end{align}
	The policy is then found by taking the actions that maximize the action-value function, usually in combination with a $\varepsilon$-greedy policy for exploration.

	\paragraph*{Deep Reinforcement Learning} 
	A key element of modern architectures is the use of neural network based function approximation to extract relevant features for value estimation.
	In this work, we focus on Deep Q-Networks (DQN) \cite{mnih_human-level_2015} and Soft Actor-Critic (SAC) \cite{haarnoja_soft_2018-1} as action-value based off-policy agents.
	Being off-policy allows to reuse collected experience for sample efficiency but causes problems if the agent should adapt to drifting conditions as the environment might behave differently than the stored experience suggests.
	In contrast to the basic $\varepsilon$-greedy used by DQN, SAC additionally uses entropy guided exploration which includes automatic entropy adjustment and thus should be more robust to the influence of drift.

	\subsection{Non-stationary Reinforcement Learning and Uncertainty} \label{sec:FOUNDATION_NSRL}
	In the broadest sense, non-stationary Reinforcement Learning is defined as an MDP where the transition dynamics as well as the reward function can change with time $t$, i.e.  $(\mathcal{S}, \mathcal{A}, P_t, R_t, \gamma)$.
	Note however, that there is large body of related work discussing different possible definitions; a non-exhaustive list includes the case of bandits \cite{liu_definition_2023}, NSRL as a special case of partially observable MDPs (POMDPs) \cite{khetarpal_towards_2022-1} or an agent centric perspective \cite{abel_definition_2023-1}.
	From a formal perspective the broad definition is sufficient for this work and we in contrast want to focus on practical constraints that we discuss in the following.
	
	In stationary RL settings it is common to scale simulations by using multiple copies of an environment, in some cases this is also transferred to real-world systems \cite{levine_learning_2017, kalashnikov_scalable_2018}.
	However, most deployed industrial systems cannot be scaled in this way: usually only a single copy of a system is available \cite{dulac-arnold_challenges_2021} rendering data collection expensive and scaling simulated copies suffers from the sim-to-real gap \cite{zhao_sim--real_2020}.
	Further, changes in industrial systems occur throughout their lifetime and can often be attributed to influences such as temperature or wear-and-tear: hence it is highly unrealistic to assume the availability of multiple environments affected by the same form of non-stationarity.
	Thus, a deployed RL usually interacts with a \emph{single system} at a \emph{single point in time}, hence this work addresses what we term SEOS-NSRL.
	\begin{assumption}[Single Environment, One-Shot Non-Stationary RL (SEOS-NSRL)]\label{NSRL_ASSUMPTION}
		An agent is only able to interact with a \textbf{single environment}.
		This environment might drift \textbf{at every timestep} a) without the knowledge of the agent, b) across episodes and c) rendering the current policy suboptimal -- \textbf{requiring} policy adaptation.
	\end{assumption}
	Drifting environments are common throughout industry.
	One example is the use of linear transport systems: such systems are ubiquitous in industrial plants and range from simpler conveyor-like systems to more complex independent cart systems.
	The latter usually rely on bearings, rollers or similar technology which are slowly affected by wear-and-tear over long periods of time \cite{cavalaglio_camargo_molano_bearing_2020}.
	Further, the wear manifests locally due to the use of multiple independent carts, each again subject to significantly different extent of usage \cite{jabbar_moira-unimore_2025}.
	Therefore, an RL agent interacting with such a system has to expect such variations at all times and separately for each affected component.
	
	In this setting, an RL agent has to face several challenges; two of those are particularly relevant to this work and address the uncertainty the agent has to handle. More precisely, uncertainty can occur in (at least) two different ways:\vspace{-5pt}
	\begin{enumerate}[\alph{enumi})]\setlength\itemsep{0em}
		\item The agent relies \textbf{on old} information: deprecated transitions are still present in the replay buffer or indirectly encoded in for example the weights of a neural network.
		\item The agent has \textbf{no} information: neural networks are not designed for extrapolation, i.e. states that are not visited might take arbitrary  values (over- or underestimation).
	\end{enumerate}\vspace{-5pt}
	The former can only occur in NSRL whereas the latter is more general: yet as shown in \Cref{fig:visabs} overestimation seems to be particularly relevant in NSRL settings.
	The goal of our novel forgetting mechanism is to enforce a baseline value $b$ in states where we do not have any or only old information which is done by sampling possible states independent of the used policy and collected experience -- we provide a biological motivation for such an approach next.
	%This is done by performing a similar update to \Cref{eq:NtVD} but assuming the possibility to \emph{sample} from the relevant action/observation spaces which we discuss in \Cref{sec:ssvd}.
	
	\subsection{Biological Motivation for Forgetting Mechanisms}
	In behavioral biology, Q-Learning with forgetting mechanisms emerged \cite{ito_validation_2009,beron_mice_2022} for simple, usually single state, environments and a set of discrete actions.
	Whilst slight variations of these mechanisms exist, their idea is to extend \cref{eq:QL} by a forgetting mechanism  as follows:
	\begin{align}\label{eq:NtVD}
		Q(s_t, a) &\gets (1-\eta)Q(s_t, a) +\eta b, \quad \forall a\neq a_t. 
	\end{align}
	That is, whilst the taken action is updated, the action values for all other actions decay towards a baseline $b$ (often $b=0$) with a forgetting rate $\eta$ as it is assumed that the information is less reliable.
	
	\section{Method: Space-sampled Value Decay}\label{sec:ssvd}
	Again considering the setting of tabular Q-Learning with discrete actions, we now need to define a distribution $\mathcal{D}_\mathcal{S}$ on the state space $\mathcal{S}$ to allow sampling.
	Terming this novel approach \emph{Space-sampled Value Decay} (SsVD) we extend \cref{eq:QL} by another update which can be written as:
	\begin{align}\label{eq:SsVD}
		Q(s', a) &\gets (1-\eta)Q(s', a) +\eta b, \quad s'\sim\mathcal{D}_\mathcal{S}, \quad \forall a\in\mathcal{A}.
	\end{align}
	Thus, we now sample the state randomly from the state-space and decay the values of all possible actions (it is possible to also sample the action which we make use of later for SAC).
	In many real world applications the observation space is known a priori, for example the workspace configuration of a robot is usually fixed (e.g. acceptable joint angles).
	Such an approach is not directly transferable to settings where no such assumption can be made, e.g. in many video games there is no direct way to sample states -- we discuss this and possible remedies in \Cref{sec_app:sampling}.
	
	\subsection{Theoretical Intuition}
	Whilst the idea of SsVD is straight-forward, this section is dedicated to theoretically analyzing if this novel update behaves as expected
	from a theoretical viewpoint.
	Assume the current timestep is $t$ and that a state-action pair $(s,a)$ has not
	been updated by the classical Q-learning update for $\tau$ timesteps. Further,
	we assume that the change of the optimal action-value function between two
	consecutive timesteps is bounded by $\Delta \ge 0$:
	\begin{equation}\label{eq:assumption}
		|Q^*_{t+1}(s,a) - Q^*_t(s,a)| \le \Delta, \quad \forall (s,a) \in \mathcal{S}\times\mathcal{A},\ \forall t \ge 0.
	\end{equation}
	
	\begin{theorem}\label{th:main_th}
		Let $\eta\in[0,1]$, fix a state-action pair $(s,a)$ and let $t_0$ denote the last time step at
		which $Q(s,a)$ was updated by the Q-learning update. For $t \ge t_0$, let
		$\tau := t - t_0$ denote the number of time steps since this update, and
		assume that the SsVD update is applied to $(s,a)$ at every time step
		$k \in \{t_0+1, \dots, t\}$. Then
		\[
		Q_t(s,a) - Q^*_t(s,a) = (1-\eta)^{\tau}\big(Q^*_{t_0}(s,a) - Q^*_t(s,a)
		+ \epsilon_{t_0}\big) + \big(1 - (1-\eta)^{\tau}\big)\big(b - Q^*_t(s,a)\big),
		\]
		where $\epsilon_{t_0} := Q_{t_0}(s,a) - Q^*_{t_0}(s,a)$ is the estimation
		error at time $t_0$.
	\end{theorem}
	\begin{proof}
		All proofs are found in Appendix \ref{sec_app:proofs}.
	\end{proof}
	\Cref{th:main_th} now allows us to analyze how we can bound the difference between estimated and optimal action-value function.
	Taking the norm and applying some simple transformations yields:
	\begin{corollary}\label{corr:norm}
		Let $\eta$, $(s,a)$, $t_0$ and $\tau$ be as in Theorem~1. Then
		\[
		\big|Q_t(s,a) - Q^*_t(s,a)\big|
		\le \big(1 - (1-\eta)^{\tau}\big)\,\big|Q^*_t(s,a) - b\big|
		+ (1-\eta)^{\tau}\big(|\epsilon_{t_0}| + \Delta\tau\big).
		\]
	\end{corollary}
	Note that we aim for this bound to be as small as possible.
	Hence, this now becomes an optimization problem with straightforward optima where we search for an optimal value of the forgetting rate $\eta$.
	\begin{corollary}\label{corr:optimal_eta}
		Let $(s,a)$, $t_0$ and $\tau \ge 1$ be as in Theorem~1, and let
		\[
		D := \big|Q^*_t(s,a) - b\big|
		\quad\text{and}\quad
		S := |\epsilon_{t_0}| + \Delta\tau .
		\]
		Then the bound in Corollary~1, viewed as a function of $\eta \in [0,1]$, is
		minimized for $\eta = 1$ if $S > D$ and for $\eta = 0$ if $S < D$.
	\end{corollary}
	The interpretation is the following: it is optimal to either completely forget ($\eta=1$) or not forget at all ($\eta =0$), in practice however, we usually do not know whether $S>D$ or $S<D$ holds (e.g. is the baseline more informative than the estimation error), hence we assume $\eta\in(0,1)$.
	In this case, when a state is not visited but regularly sampled using SsVD (i.e. $\lim_{\tau\to\infty}$) we as intended have:
	\begin{corollary}\label{eq:op_init_val}
		Let $\eta \in (0,1)$ and let $(s,a)$ be a state-action pair that was last
		updated by the classical Q-learning update at time $t_0$, i.e.\ it receives
		no Q-learning update for all $t > t_0$, while the SsVD update is applied in
		every timestep. Then, with $\tau = t - t_0$,
		\[
		\lim_{t \to \infty} Q_t(s,a) = b.
		\]
	\end{corollary}
	i.e. a state not updated by an agent (for example it is not visited) assumes the baseline value.
	If this baseline value is chosen roughly neutral, e.g. $b\approx 0$\footnote{What counts as \emph{neutral} is environment and scale-dependent.}, this imposes a neutrality bias on states the agent is uncertain.
	A different mode of operation can be achieved by setting an optimistic prior by choosing $b$ optimistically, e.g. $b\geq V_{\max}$, which drives the agent to explore.

	\subsection{Extension to Function Approximation} \label{sec:main_method}
	Note that SsVD \emph{does not} rely on the associated rewards of the actions, i.e. it is sufficient to sample states for DQN and actions \emph{and} states for continuous actor-critic models.
	Adapting the forgetting mechanisms for modern RL algorithms is done by adding an additional auxiliary loss to optimize.
	
	\paragraph*{DQN} Deep Q-networks rely on fixing the target networks used in the bootstrapping step, we denote the frozen network as $Q^-$ \cite{mnih_human-level_2015}. 
	If the default DQN loss is given by $\mathcal{L}_{DQN}$ based on a mini-batch of size $m$ we additionally sample $p\leq m$ states $\hat{s}_1,...,\hat{s}_p$ and compute
	\begin{align}
		\mathcal{L}_F = \frac{1}{p}\sum_{i=1}^{p} \lVert Q(\hat{s}_i, \cdot) - (1-\eta) Q^-(\hat{s}_i, \cdot) - \eta \mathbf{b}\rVert ^2
	\end{align}
	with baseline $\mathbf{b}\in\mathbb{R}^{|\mathcal{A}|}$ written as a vector.
	Note that we write $Q(\hat{s}_i, \cdot)$ to denote that the action-value in DQN is implemented as $Q:\mathcal{S}\to \mathbb{R}^{|\mathcal{A}|}$ where $Q(s,a)$ is computed by picking the corresponding entry in the resulting function output which is possible as $\mathcal{A}$ is discrete and finite.
	Thus, the final loss is given by $\mathcal{L}=\mathcal{L}_F + \mathcal{L}_{DQN}$.
	In practice, we choose a fraction $\xi$ of the original mini-batch size as 	$p=\max(\lceil\xi m\rceil, 1)$, i.e. the ``max'' always guarantees that at least a single sample is used.

	\paragraph*{SAC}
	A key difference to DQN is that the critic network is implemented as $Q:\mathcal{S}\times \mathcal{A} \to \mathbb{R}$  which together with a separate actor/policy network allows the extension to continuous action spaces \cite{haarnoja_soft_2018-1}.
	In contrast to DQN we now sample $p\leq m$ states \emph{and} actions $\hat{s}_1,...,\hat{s}_p,\hat{a}_1,...,\hat{a}_p$ to compute the loss as:
	\begin{align}
		\mathcal{L}_F = \frac{1}{p}\sum_{i=1}^{p} \left[Q_j(\hat{s}_i, \hat{a}_i) - (1-\eta) Q_j^-(\hat{s}_i, \hat{a}_i) - \eta b\right] ^2
	\end{align}
	for $j=1,...,k$ with $Q_j$ representing one of the $k$ (usually $k=2$) networks as SAC uses multiple critics \cite{fujimoto_addressing_2018} to avoid overestimation, and $b$ being a scalar baseline value.
	The number of samples $p$ is also chosen according to $p=\max(\lceil\xi m\rceil, 1)$.

	\paragraph*{Sampling} Sampling complex state spaces (such as images) is generally difficult: in most cases this will require problem specific adaptations.
	Environments used in this work allow to generate arbitrary environment configurations which we can generate on the fly for sampling -- for details see \Cref{sec_app:sampling}. 
	A similar limitation applies to higher dimensional spaces which are difficult to sample due to the curse of dimensionality and if it is unclear which states actually represent realistic environment configurations (e.g. some dimensions might allow theoretically unbounded values).
	
	\begin{table*}[t!]
		\centering
		\caption{Overview of used environments grouped by type. Stationary environments to evaluate SsVD for exploration are marked in purple and by a $\dagger$ symbol.}
		\label{tab:drift_environments}
		\small
		\resizebox{\textwidth}{!}{
		\begin{tabular}{cllllc}
			\toprule
			& \textbf{Environment} & \textbf{Non-st. Parameter} & \textbf{Type of Drift} & \textbf{Scheduler} & \textbf{Actions}\\
			\midrule
			\multirow{3}{*}{\rotatebox[origin=c]{90}{Classic}}
			& CartPole-v1 & Force & Decr. ($-$) & Probab.  & Discrete \\
			& MountainCar-v0 & Force & Decr. ($-$) & Probab.  & Discrete \\
			& \spec{ MountainCarContinuous-v0} & \emph{None} & \emph{None} & \emph{None}  & Continuous \\
			\midrule
			\multirow{1}{*}{\rotatebox[origin=c]{90}{\small MJ}}
			& InvertedPendulum-v5 & Cart mass & Decr. ($-$) & Probab. & Continuous \\
			\midrule
			\multirow{4}{*}{\rotatebox[origin=c]{90}{\small Misc.}}
			& \spec{DeepSea} & \emph{None} & \emph{None} & \emph{None} & Discrete  \\
			& \spec{RiverSwim} & \emph{None} & \emph{None} & \emph{None} & Discrete  \\
			& MultiPadLunarLander (cont./discr.) & Reward & Flip (la. pad) & Probab. & Discrete/Continuous  \\
			& ChooseBox & Reward & Flip ($-1\cdot$) & Fixed  & Discrete  \\
			\bottomrule
		\end{tabular}
	}
	\end{table*}
	
	\section{Empirical Experiments}\label{sec:exps}
	The code used for all experiments and to reproduce the shown results exactly is publicly available on GitHub\footnote{\ifanonym
			\url{https://anonymous.4open.science/r/SpaceSampledValueDecay-3F20}
		\else
			\url{https://github.com/Felix-St/SpaceSampledValueDecay}
		\fi}.
	Further details can be found in Appendix \ref{sec:app_add_det}.
	
	\subsection{Non-stationary Environments}\label{sec:emp_ana_nsrl}
	
	\paragraph*{Algorithms} 
	We run our proposed SsVD forgetting mechanism on top of the base algorithms DQN and SAC (we denote these extensions by -F, i.e. DQN-F for Deep Q-Networks with SsVD).
	Further, we include limited (i.e. non-tracking) versions of the base algorithms to illustrate what happens if no further adaptation is done.	
	As an on-policy tracking baseline \cite{sutton_role_2007}, we include PPO \cite{schulman_proximal_2017} restricted to a single environment instance to strictly adhere to the SEOS-NSRL setting (\Cref{NSRL_ASSUMPTION}), alongside off-policy tracking baselines (DQN and SAC).
	Further, we compare against two general purpose baselines; \citet{nikishin_primacy_2022} use recurring resets of the network parameters while maintaining the replay buffer and one with L2 regularization towards the initial parameters of the network \cite{kumar_maintaining_2025}.
	All implementations rely on Stable Baselines 3 \cite{raffin_stable-baselines3_2021} using the \emph{exact same} (non method-specific) parameters adapted to the SEOS-NSRL setting (i.e. we use a single environment throughout).
	The used parameters correspond to the pre-tuned values found in RL zoo \cite{raffin_rl_2020}: further details can be found in \Cref{sec_app:hpo}.
	
	\begin{figure*}[t!]
		\centering
		\includegraphics[width=1\textwidth]{
			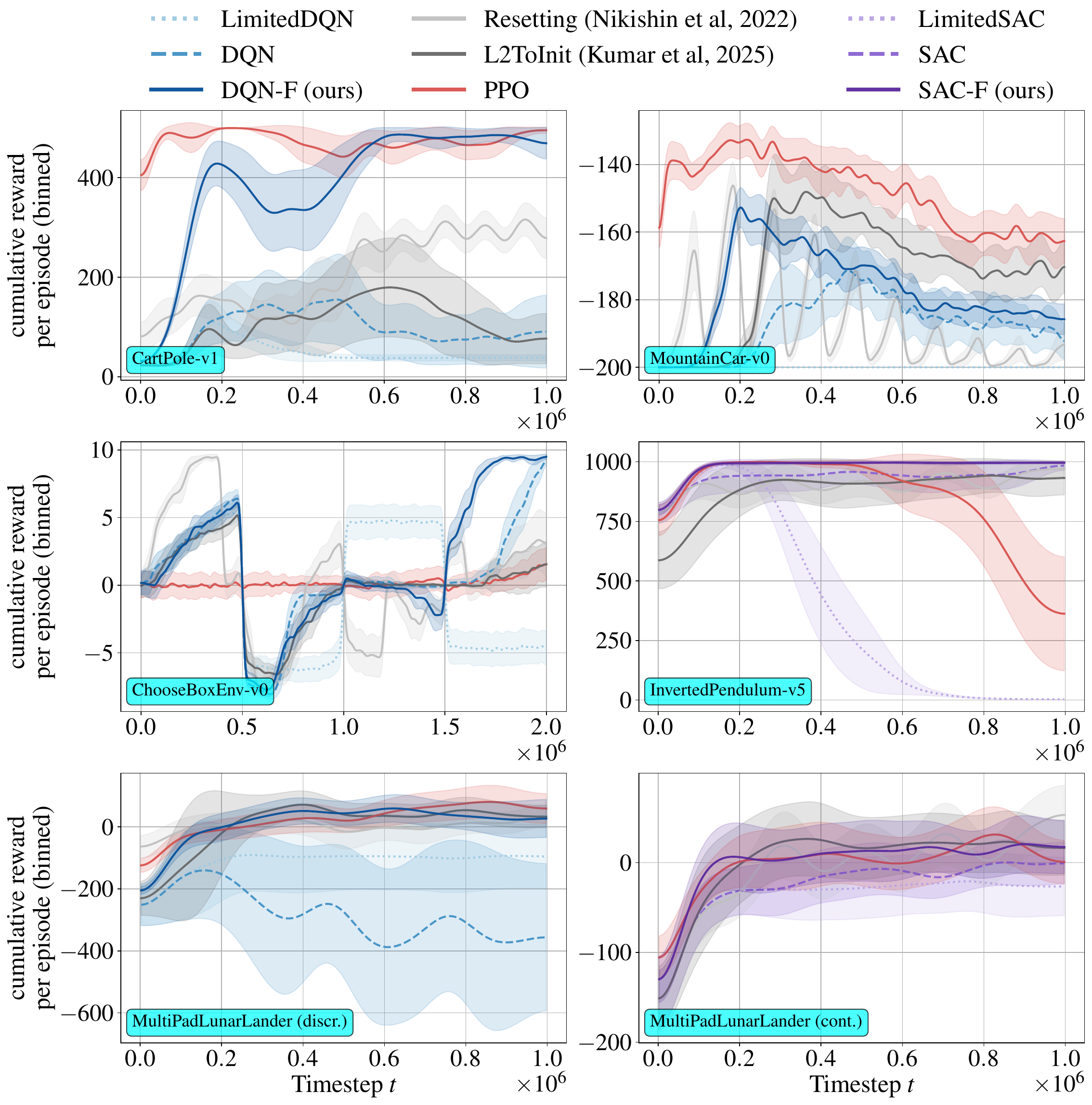}
		\caption{Empirical evaluation on $6$ non-stationary environments.
			Smoothed cumulative returns of terminating episodes mean-aggregated across bins (y-axis, solid lines) with $0.5\sigma$ empirical standard deviation (shaded).
			Timesteps indicates the number of environment interactions (x-axis).}
		\label{fig:main_result}
	\end{figure*}
	
	\paragraph*{Environments and Drifts}
	We use the classic control environments \env{CartPole} and \env{MountainCar} as well as \env{InvertedPendulum} as a MuJoCo based environments.
	For these environments, we utilize \citet{keplinger_ns-gym_2025-1} to define continuous and incremental drifts that modify a parameter of the dynamics (e.g. the power of the car in \env{MountainCar} as indicated in \Cref{fig:visabs}).
	We additionally introduce two novel non-stationary environments with flipping rewards as drifting dynamics: \env{MultiPadLunarLander}, an extension of the classic \env{LunarLander} environment with a second landing pad and \env{ChooseBox}, a Minigrid \cite{chevalier-boisvert_minigrid_2023} based environment with the goal of picking up one of two boxes, for details see \Cref{sec:custom_envs}.
	Only \env{ChooseBox} uses a number of fixed timesteps where the drift occurs -- all other drifts are stochastic as the drift is applied with a given probability (these thus vary between seeds) -- details on this are found in \Cref{sec_app:non_st_ex}.
	\Cref{tab:drift_environments} provides a full list of environments used herein. 
	
	\paragraph*{Evaluation}
	Episode rewards are mean aggregated across timesteps using fixed-width bins of size $B= 5 T_{\max}$ where $T_{max}$ is the maximum episode length. For each random seed (we run 10 iterations on each environment), rewards within a bin are averaged, and the mean and standard deviation are then computed across seeds centered at each bin. 
	Number of runs (i.e. seeds) per environment are listed in \Cref{tab:hyperparams}.
	We further apply Gaussian smoothing with $\sigma_{\text{smooth}}=10$ to both.
	
\begin{figure*}[t!]
		\centering
		\includegraphics[width=1\textwidth]{
			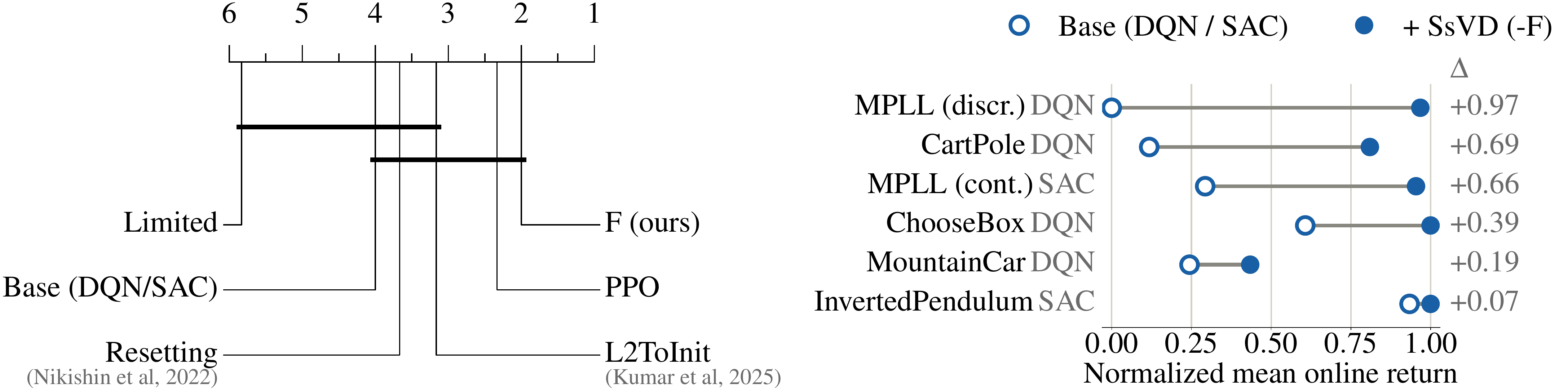}
		\caption{\textbf{Left:} Mean ranking (lower is better)  of the algorithms within-environment ranks using Autorank \cite{herbold_autorank_2020}. The within-environment rank is computed based on the mean online return. The black-bar marks models where the ranking cannot be guaranteed with statistical significance. \textbf{Right:} Comparison of the normalized (across all methods) mean online return (higher is better) between the base variant and its combination with SsVD across all environments.}
		\label{fig:ranking}
	\end{figure*}

	\paragraph*{Results}
	\Cref{fig:main_result} shows the online return during training; the mean online return is listed in \Cref{tab_app:mor} (Appendix \ref{sec_app:add_res}).
	Of the base algorithms, DQN fails to adapt likely because its $\varepsilon$-schedule is not designed for drifting environments.
	Adding SsVD clearly improves over the respective base algorithm on all six environments (\Cref{fig:ranking}, right).
	The -F variants also attain a higher mean online return than Resetting on all six environments and than L2ToInit on four; L2ToInit is ahead on \env{MountainCar} and on both \env{MultiPadLunarLander} variants the differences between all three methods are small.
	PPO is a strong tracking baseline since, being on-policy, it is not affected by outdated transitions: it attains the highest mean online return on \env{CartPole}, \env{MountainCar} and \env{MultiPadLunarLander (discr.)} and is on par with SAC-F on \env{MultiPadLunarLander (cont.)}.
	However, PPO fails to learn \env{ChooseBox} and its return on \env{InvertedPendulum} drops sharply late in training.
	Thus, neither approach dominates.
	Aggregated over all environments, the -F variants attain the best mean rank (closely followed by PPO, \Cref{fig:ranking}, left), although this is not statistically significant \cite{herbold_autorank_2020}.
	
	\paragraph*{Stationary Ablation} We rerun part of the same experiment with stationary environments -- a setting where our method also improves on the base algorithms, see Appendix \ref{sec_app:add_res}.
	
	\subsection{Qualitative Analysis on MountainCar}\label{sec:qualitative}
	For a qualitative, visual analysis we choose the \emph{discrete} non-stationary \env{MountainCar} environment as its state space is two-dimensional; we visualize the learned value surface ($\max_a Q(s, a)$) by taking snapshots at intermediate timesteps and sampling rollout trajectories.
	Note that despite the problem's age and apparent simplicity, it has been shown by \citet{huber_chebyshev_2026} that modern algorithms only achieve suboptimal regret on the stationary (albeit continuous) variant of this task.
	
	From \Cref{fig:visabs} it is evident that vanilla DQN starts to drastically overestimate the values associated around the goal state in the top right.
	Weight decay is enabled here and does not prevent this.
	As classic Q-learning is known to be overestimating \cite{hasselt_double_2010} we compare with DoubleDQN \cite{hasselt_deep_2016} which is aimed at mitigating overestimation and also an ablation using a bounded activation function instead of the ReLU default.
	\Cref{fig_app:doubledqn} (Appendix \ref{sec_app:add_res}) clearly indicates that using DoubleDQN already reduces but not removes the issue of overestimation -- however when using a bounded activation function this issue does not occur.
	Hence, we believe that the overestimation is caused by the continued bootstrapping over longer-than-usual time horizons.
	
	In DQN-F, the region usually visited (centered diamond-like shape) is well recognizable.
	Values that are still in-distribution in principle but outside this typical region are pulled towards the $b=0$ baseline which thus form a white colored \emph{ring} around it -- note that $b=0$ is an \emph{optimistic} baseline for this environment and that weight decay is \emph{not} used.
	This is \Cref{eq:op_init_val} realized empirically as the ring is a mild exploration bonus surrounding the current behavior: as the power decreases and the previously successful swing pattern stops reaching $x > 0.5$, these surrounding states become comparatively attractive and re-exploration follows. 
	This is also evident from sampled policy rollouts: the trajectories lengthen as more swings are required but in nearly all cases reach the goal state.
	
	\subsection{Hard Exploration}\label{sec:hard_exploration}
	\paragraph*{Setup} A different interpretation of SsVD is possible by choosing optimistic values for the baseline $b$ to aid in exploration.
	To evaluate this additional property we choose a set of three standard benchmark environments for hard exploration tasks (see \Cref{tab:drift_environments}, marked) whilst the environments are kept stationary and the baseline value is initialized optimistically.
	As commonly done with exploration heuristics, we start with a large bonus ($b\propto V_{\max}$) which we linearly decay (either $b$, $\xi$ or $\eta$) over roughly the first half of training, enabling asymptotic policy convergence (see \Cref{sec_app:hpo}).
	
	For \env{MountainCarContinuous} and \env{RiverSwim} we perform the same evaluation protocol as before but for \env{DeepSea} (details on the latter two environments can be found in \Cref{sec:exploration_envs}) we scale the problem complexity (i.e. grid size), following \citet{osband_deep_2019} we include BootstrappedDQN with random priors as a baseline.	
	
\begin{figure*}[t!]
	\centering
	\includegraphics[width=1\textwidth]{
		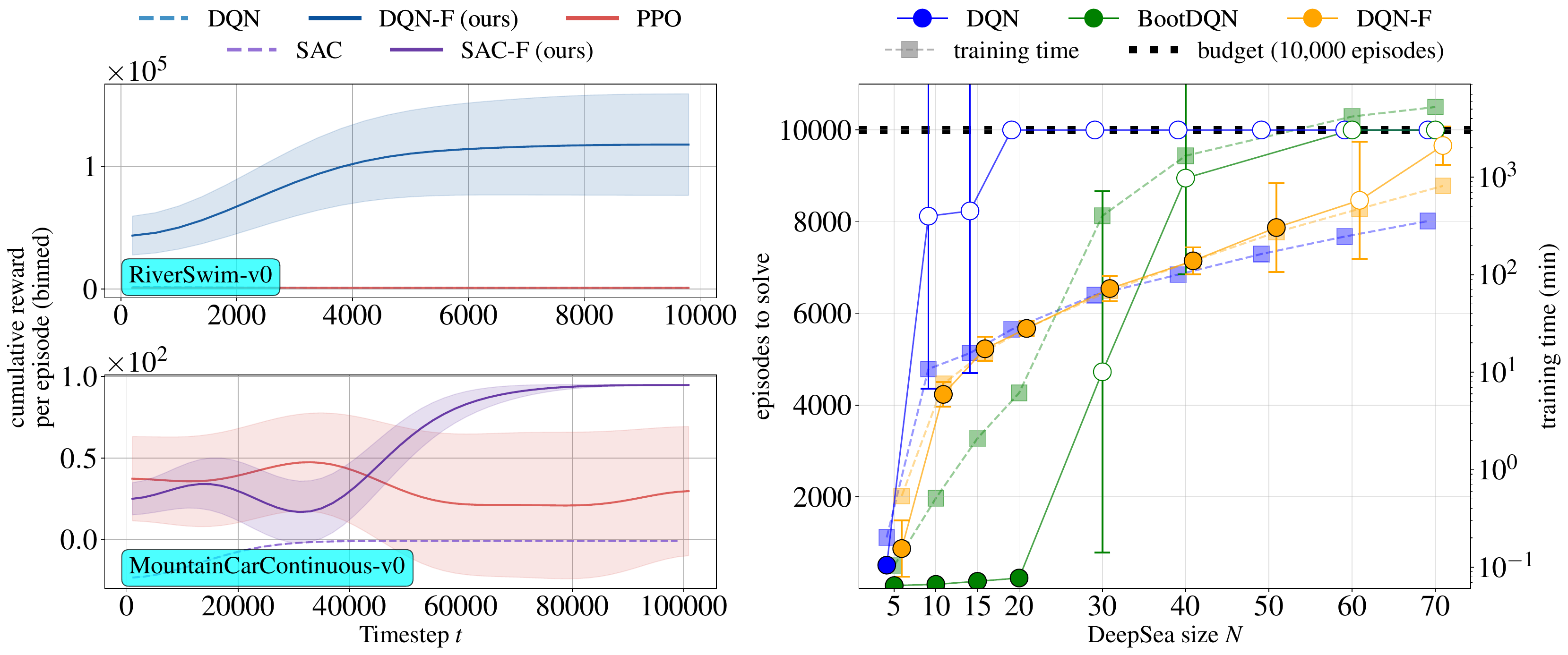}
	\caption{SsVD evaluated on hard exploration tasks. \textbf{Left:} SsVD with an optimistic baseline $b$ on 2 stationary exploration environments. Otherwise same evaluation setup as in \Cref{fig:main_result} for $10$ runs with $\sigma_{\text{smooth}}=5$.
		\textbf{Right:} Exploration on \env{DeepSea} for different grid sizes (x-axis) and the mean number of episodes required to solve (y-axis, left) as well as wall-clock time required (y-axis, right) across 5 seeds (if a seed is unsolved the marker is hollow and counts as the maximum budget).
	}
	\label{fig:hard_exploration}
\end{figure*}
	
	\paragraph*{Discussion} \Cref{fig:hard_exploration} clearly shows that SsVD with positive baseline values is beneficial for exploration.
	Note that the use of generalized State Dependent Exploration (gSDE) \cite{raffin_smooth_2022} is disabled on \env{MountainCarContinuous} to isolate the effect of using SsVD. 
	On \env{DeepSea} we can see that our proposed method reliably solves the environment up to $N=50$ and occasionally up to $N=70$; outperforming the BootstrappedDQN baseline for $N\geq30$.

	\section{Related Work}\label{sec:rew_work}
	
	\paragraph*{Planning and Trajectory Sampling} 
	Classical approaches rely on sampling states for model-based planning (e.g. trajectory sampling) or exploring starts \cite{sutton_reinforcement_2018} and reward baselines are widely used to reduce variance \cite{williams_simple_1992,baird_advantage_1993,weaver_optimal_2001}; SsVD however works \emph{independent} of a) visiting a state (though choosing visited states would be possible) and b) of the actual dynamics of the environment as SsVD does not aim to estimate their values.

	\paragraph*{Non-Stationary / Continual RL} A large body of literature focuses on a setting of discrete tasks or sets of change points \cite{kirkpatrick_overcoming_2017,al-shedivat_continuous_2018,rolnick_experience_2019, kauvar_curious_2023,caccia_task-agnostic_2023, ahn_prevalence_2025}, some also assuming intra episode change that resets \cite{keplinger_ns-gym_2025-1}.
	Many (though not all e.g. \citealp{caccia_task-agnostic_2023}) also assume additional information such as context or task IDs. Such a setting is also referred to as Life-long RL \cite{abel_policy_2018} -- usually an additional goal is to mitigate forgetting of previous knowledge.
	These assumptions are however unrealistic for real-world applications \cite{behdin_position_2026}.
	In contrast, we here take the more realistic viewpoint of SEOS-NSRL (see \Cref{NSRL_ASSUMPTION}), requiring that old information has to be forgotten to maintain optimality.
	To the best of the authors' knowledge, only a few works explicitly address this view: \citet{xie_deep_2021, liu_learning_2025} both rely on learning latent-variables to infer the current or next state the environment is in; \cite{liu_learning_2025} also uses change-points detection.
	Others rely on more general strategies that work independent of the setting such as periodic resets \cite{nikishin_primacy_2022} or modified L2 regularization \cite{kumar_maintaining_2025}.

	\paragraph{Exploration and Optimism}
	When parametrized with an optimistic baseline (see also \Cref{eq:op_init_val}), SsVD yields a variant of optimistic initialization \cite{marlos_c_machado_domain-independent_2015, sutton_reinforcement_2018} for function approximation.
	While visitation-count methods \citep{bellemare_unifying_2016,sutton_reinforcement_2018} and curiosity mechanisms like RND \citep{burda_exploration_2019} compute intrinsic rewards along visited or simulated trajectories, SsVD operates directly on unvisited state representations, driving exploration toward unexplored regions without requiring environment transitions.

	\section{Conclusion, Limitations, and Future Work}
	Biologically inspired, we introduced \emph{Space-sampled Value Decay} (SsVD) as a simple, effective forgetting mechanism for deep off-policy RL agents (DQN and SAC). 
	SsVD stabilizes learning in diverse settings: from non-stationary adaptation to mitigating overestimation and driving exploration.
	
	\paragraph*{Limitations} 
	Non-stationarity in physical systems manifests heterogeneously, and no single adaptation strategy is optimal across all drift profiles without prior knowledge. 
	A primary practical limitation of SsVD is its reliance on sampling valid environment states, which becomes non-trivial in high-dimensional or image-based observation spaces without world models or generative priors. 
	Additionally, while the baseline $b$ can be chosen based on task return scales, SsVD introduces auxiliary hyperparameters ($\eta$ and $\xi$) that may require tuning depending on rate and magnitude of drift.
	
	\paragraph*{Future Work} 
	A natural next step is to explore more adaptive sampling strategies that focus forgetting on states where action-value uncertainty is highest, bypassing uniform sampling in complex state spaces. 
	This encompasses both domain-specific geometric schemes \citep[e.g.,][]{yershova_deterministic_2004} but also task-agnostic representations. 
	Finally, developing adaptive mechanisms that dynamically tune the forgetting rate $\eta$ could further be beneficial for autonomously deploying SsVD.

	\subsection*{AI use statement}
	We used generative AI tools for finding related and relevant work (deep research) and also as a coding agent to support in the development of software; all generated code and found related work were manually checked for correctness. 
	Further, an extensive manual survey of the literature was performed.
	The elements of the theory section (i.e. theorems and proofs) were manually drafted and then generative AI was applied to clean and improve their presentation.
	Generative AI was \emph{not} used for writing or contributing to the main text of the paper except for grammar checking, minor rephrasing and providing high-level feedback on the general structure of the paper. Generative AI helped in writing parts of the text found in the Appendix.
	
	\subsection*{Reproducibility Statement}
	Full code is made available on GitHub (see \Cref{sec:exps}).
	The code not only includes all implementations of benchmarked algorithms including the environments, but also all files used to run the empirical evaluation and to exactly generate the herein shown plots.
	Further, results underlying the quantitative analysis are provided as ``.json'' files.
	A requirements file is provided for installing the used software libraries and to document their used versions.
	Used parameters are not only found in the provided code but also listed in \Cref{tab:hyperparams}.
	The two newly introduced environments are described in more detail in \Cref{sec:custom_envs} and the introduced forms of non-stationarities are also shown visually and discussed in detail in \Cref{sec_app:non_st_ex}.
	
	\ifanonym
	
	\else
		\subsubsection*{Acknowledgements} 
		Funding in the scope of the ERC Synergy Grant ``Water-Futures'' No. 951424 is gratefully acknowledged.\\
		
		This work was supported by the Fraunhofer Internal Programs under
		Grant No. SME 40-09551
	\fi
	
	\bibliography{iclr2027_conference}

@inproceedings{abel_definition_2023-1,
  title = {A Definition of Continual Reinforcement Learning},
  booktitle = {Advances in Neural Information Processing Systems},
  author = {Abel, David and Barreto, Andre and Van Roy, Benjamin and Precup, Doina and {van Hasselt}, Hado and Singh, Satinder},
  editor = {Oh, A. and Naumann, T. and Globerson, A. and Saenko, K. and Hardt, M. and Levine, S.},
  year = 2023,
  volume = {36},
  pages = {50377--50407},
  publisher = {Curran Associates, Inc.}
}

@inproceedings{abel_policy_2018,
  title = {Policy and Value Transfer in Lifelong Reinforcement Learning},
  booktitle = {Proceedings of the 35th International Conference on Machine Learning},
  author = {Abel, David and Jinnai, Yuu and Guo, Sophie Yue and Konidaris, George and Littman, Michael},
  editor = {Dy, Jennifer and Krause, Andreas},
  year = 2018,
  month = jul,
  series = {Proceedings of Machine Learning Research},
  volume = {80},
  pages = {20--29},
  publisher = {PMLR}
}

@inproceedings{ahn_prevalence_2025,
  title = {Prevalence of Negative Transfer in Continual Reinforcement Learning: {{Analyses}} and a Simple Baseline},
  booktitle = {International Conference on Learning Representations},
  author = {Ahn, Hongjoon and Hyeon, Jinu and Oh, Youngmin and Hwang, Bosun and Moon, Taesup},
  editor = {Yue, Y. and Garg, A. and Peng, N. and Sha, F. and Yu, R.},
  year = 2025,
  volume = {2025},
  pages = {75036--75060}
}

@inproceedings{al-shedivat_continuous_2018,
  title = {Continuous {{Adaptation}} via {{Meta-Learning}} in {{Nonstationary}} and {{Competitive Environments}}},
  booktitle = {International {{Conference}} on {{Learning Representations}}},
  author = {{Al-Shedivat}, Maruan and Bansal, Trapit and Burda, Yura and Sutskever, Ilya and Mordatch, Igor and Abbeel, Pieter},
  year = 2018,
  month = feb,
  urldate = {2026-08-31},
  langid = {english}
}

@techreport{baird_advantage_1993,
  type = {Technical {{Report}}},
  title = {Advantage Updating},
  author = {Baird, III, Leemon C.},
  year = 1993,
  number = {WL-TR-93-1146},
  address = {Dayton, OH},
  institution = {Wright Laboratory, Wright-Patterson Air Force Base}
}

@inproceedings{behdin_position_2026,
  title = {Position: {{Deployed Reinforcement Learning}} Should Be {{Continual}}},
  booktitle = {Forty-{{Third International Conference}} on {{Machine Learning}}},
  author = {Behdin, Parnian and Roice, Kevin and Mesbahi, Golnaz},
  year = 2026,
  langid = {english}
}

@inproceedings{bellemare_unifying_2016,
  title = {Unifying Count-Based Exploration and Intrinsic Motivation},
  booktitle = {Advances in Neural Information Processing Systems},
  author = {Bellemare, Marc and Srinivasan, Sriram and Ostrovski, Georg and Schaul, Tom and Saxton, David and Munos, Remi},
  editor = {Lee, D. and Sugiyama, M. and Luxburg, U. and Guyon, I. and Garnett, R.},
  year = 2016,
  volume = {29},
  publisher = {Curran Associates, Inc.}
}

@article{beron_mice_2022,
  title = {Mice Exhibit Stochastic and Efficient Action Switching during Probabilistic Decision Making},
  author = {Beron, Celia C. and Neufeld, Shay Q. and Linderman, Scott W. and Sabatini, Bernardo L.},
  year = 2022,
  month = apr,
  journal = {Proceedings of the National Academy of Sciences},
  volume = {119},
  number = {15},
  pages = {e2113961119},
  publisher = {Proceedings of the National Academy of Sciences},
  urldate = {2025-09-22}
}

@inproceedings{burda_exploration_2019,
  title = {Exploration by Random Network Distillation},
  booktitle = {International Conference on Learning Representations},
  author = {Burda, Yuri and Edwards, Harrison and Storkey, Amos and Klimov, Oleg},
  year = 2019
}

@inproceedings{caccia_task-agnostic_2023,
  title = {Task-Agnostic Continual Reinforcement Learning: {{Gaining}} Insights and Overcoming Challenges},
  booktitle = {Proceedings of the 2nd Conference on Lifelong Learning Agents},
  author = {Caccia, Massimo and Mueller, Jonas and Kim, Taesup and Charlin, Laurent and Fakoor, Rasool},
  editor = {Chandar, Sarath and Pascanu, Razvan and Sedghi, Hanie and Precup, Doina},
  year = 2023,
  month = aug,
  series = {Proceedings of Machine Learning Research},
  volume = {232},
  pages = {89--119},
  publisher = {PMLR}
}

@article{cavalaglio_camargo_molano_bearing_2020,
  title = {A Bearing Fault Model for {{Independent Cart Conveyor System}} and Its Validation},
  author = {Cavalaglio Camargo Molano, Jacopo and Capelli, Luca and Rubini, Riccardo and Borghi, Davide and Cocconcelli, Marco},
  year = 2020,
  journal = {Applied Acoustics},
  volume = {159},
  pages = {107069},
  issn = {0003-682X}
}

@inproceedings{chevalier-boisvert_minigrid_2023,
  title = {Minigrid \& Miniworld: Modular \& Customizable Reinforcement Learning Environments for Goal-Oriented Tasks},
  booktitle = {Proceedings of the 37th International Conference on Neural Information Processing Systems},
  author = {{Chevalier-Boisvert}, Maxime and Dai, Bolun and Towers, Mark and {de Lazcano}, Rodrigo and Willems, Lucas and Lahlou, Salem and Pal, Suman and Castro, Pablo Samuel and Terry, Jordan},
  year = 2023,
  series = {Nips '23},
  publisher = {Curran Associates Inc.},
  address = {New Orleans, LA, USA}
}

@article{dulac-arnold_challenges_2021,
  title = {Challenges of Real-World Reinforcement Learning: Definitions, Benchmarks and Analysis},
  author = {{Dulac-Arnold}, Gabriel and Levine, Nir and Mankowitz, Daniel J. and Li, Jerry and Paduraru, Cosmin and Gowal, Sven and Hester, Todd},
  year = 2021,
  month = sep,
  journal = {Machine Learning},
  volume = {110},
  number = {9},
  pages = {2419--2468},
  issn = {1573-0565},
  urldate = {2026-09-04},
  langid = {english}
}

@inproceedings{fujimoto_addressing_2018,
  title = {Addressing Function Approximation Error in Actor-Critic Methods},
  booktitle = {Proceedings of the 35th International Conference on Machine Learning},
  author = {Fujimoto, Scott and {van Hoof}, Herke and Meger, David},
  editor = {Dy, Jennifer and Krause, Andreas},
  year = 2018,
  month = jul,
  series = {Proceedings of Machine Learning Research},
  volume = {80},
  pages = {1587--1596},
  publisher = {PMLR}
}

@inproceedings{haarnoja_soft_2018-1,
  title = {Soft Actor-Critic: {{Off-policy}} Maximum Entropy Deep Reinforcement Learning with a Stochastic Actor},
  booktitle = {Proceedings of the 35th International Conference on Machine Learning},
  author = {Haarnoja, Tuomas and Zhou, Aurick and Abbeel, Pieter and Levine, Sergey},
  editor = {Dy, Jennifer and Krause, Andreas},
  year = 2018,
  month = jul,
  series = {Proceedings of Machine Learning Research},
  volume = {80},
  pages = {1861--1870},
  publisher = {PMLR}
}

@article{hasselt_deep_2016,
  title = {Deep {{Reinforcement Learning}} with {{Double Q-Learning}}},
  author = {{van Hasselt}, Hado and Guez, Arthur and Silver, David},
  year = 2016,
  month = mar,
  journal = {Proceedings of the AAAI Conference on Artificial Intelligence},
  volume = {30},
  number = {1},
  issn = {2374-3468},
  urldate = {2026-09-09},
  copyright = {Copyright (c)},
  langid = {english}
}

@inproceedings{hasselt_double_2010,
  title = {Double Q-Learning},
  booktitle = {Advances in Neural Information Processing Systems},
  author = {{van Hasselt}, Hado},
  editor = {Lafferty, J. and Williams, C. and {Shawe-Taylor}, J. and Zemel, R. and Culotta, A.},
  year = 2010,
  volume = {23},
  publisher = {Curran Associates, Inc.}
}

@article{herbold_autorank_2020,
  title = {Autorank: {{A Python}} Package for Automated Ranking of Classifiers},
  author = {Herbold, Steffen},
  year = 2020,
  journal = {Journal of Open Source Software},
  volume = {5},
  number = {48},
  pages = {2173},
  publisher = {The Open Journal}
}

@article{hochreiter_long_1997,
  title = {Long Short-Term Memory},
  author = {Hochreiter, Sepp and Schmidhuber, J{\"u}rgen},
  year = 1997,
  journal = {Neural Computation},
  volume = {9},
  number = {8},
  pages = {1735--1780}
}

@inproceedings{huber_chebyshev_2026,
  title = {Chebyshev {{Policies}} and the {{Mountain Car Problem}}: {{Reinforcement Learning}} for {{Low-Dimensional Control Tasks}}},
  shorttitle = {Chebyshev {{Policies}} and the {{Mountain Car Problem}}},
  booktitle = {Forty-Third {{International Conference}} on {{Machine Learning}}},
  author = {Huber, Stefan and Unger, Hannes and Sch{\"a}fer, Georg and Rehrl, Jakob},
  year = 2026,
  month = jun,
  urldate = {2026-09-10},
  langid = {english}
}

@article{ito_validation_2009,
  title = {Validation of {{Decision-Making Models}} and {{Analysis}} of {{Decision Variables}} in the {{Rat Basal Ganglia}}},
  author = {Ito, Makoto and Doya, Kenji},
  year = 2009,
  month = aug,
  journal = {The Journal of Neuroscience},
  volume = {29},
  number = {31},
  pages = {9861--9874},
  issn = {0270-6474, 1529-2401},
  urldate = {2025-09-22},
  copyright = {https://creativecommons.org/licenses/by-nc-sa/4.0/},
  langid = {english}
}

@article{jabbar_moira-unimore_2025,
  title = {{{MOIRA-UNIMORE}} Bearing Data Set for Independent Cart Systems},
  author = {Jabbar, Abdul and Cocconcelli, Marco and D'Elia, Gianluca and Borghi, Davide and Capelli, Luca and Cavalaglio Camargo Molano, Jacopo and Strozzi, Matteo and Rubini, Riccardo},
  year = 2025,
  journal = {Applied Sciences},
  volume = {15},
  number = {3691},
  issn = {2076-3417}
}

@inproceedings{kalashnikov_scalable_2018,
  title = {Scalable Deep Reinforcement Learning for Vision-Based Robotic Manipulation},
  booktitle = {Proceedings of the 2nd Conference on Robot Learning},
  author = {Kalashnikov, Dmitry and Irpan, Alex and Pastor, Peter and Ibarz, Julian and Herzog, Alexander and Jang, Eric and Quillen, Deirdre and Holly, Ethan and Kalakrishnan, Mrinal and Vanhoucke, Vincent and Levine, Sergey},
  editor = {Billard, Aude and Dragan, Anca and Peters, Jan and Morimoto, Jun},
  year = 2018,
  month = oct,
  series = {Proceedings of Machine Learning Research},
  volume = {87},
  pages = {651--673},
  publisher = {PMLR}
}

@inproceedings{kauvar_curious_2023,
  title = {Curious Replay for Model-Based Adaptation},
  booktitle = {Proceedings of the 40th International Conference on Machine Learning},
  author = {Kauvar, Isaac and Doyle, Chris and Zhou, Linqi and Haber, Nick},
  editor = {Krause, Andreas and Brunskill, Emma and Cho, Kyunghyun and Engelhardt, Barbara and Sabato, Sivan and Scarlett, Jonathan},
  year = 2023,
  month = jul,
  series = {Proceedings of Machine Learning Research},
  volume = {202},
  pages = {16018--16048},
  publisher = {PMLR}
}

@inproceedings{keplinger_ns-gym_2025-1,
  title = {{{NS-gym}}: A Comprehensive and Open-Source Simulation Framework for Non-Stationary Markov Decision Processes},
  booktitle = {Advances in Neural Information Processing Systems},
  author = {Keplinger, Nathaniel S. and Luo, Baiting and Zhang, Yunuo and Wray, Kyle H and Laszka, Aron and Dubey, Abhishek and Mukhopadhyay, Ayan},
  editor = {Belgrave, D. and Zhang, C. and Lin, H. and Pascanu, R. and Koniusz, P. and Ghassemi, M. and Chen, N.},
  year = 2025,
  volume = {38, Main Conference},
  publisher = {Curran Associates, Inc.}
}

@article{khetarpal_towards_2022-1,
  title = {Towards {{Continual Reinforcement Learning}}: {{A Review}} and {{Perspectives}}},
  shorttitle = {Towards {{Continual Reinforcement Learning}}},
  author = {Khetarpal, Khimya and Riemer, Matthew and Rish, Irina and Precup, Doina},
  year = 2022,
  month = dec,
  journal = {Journal of Artificial Intelligence Research},
  volume = {75},
  pages = {1401--1476},
  issn = {1076-9757},
  urldate = {2026-08-20},
  copyright = {Copyright (c) 2022 Journal of Artificial Intelligence Research},
  langid = {english}
}

@article{kirkpatrick_overcoming_2017,
  title = {Overcoming Catastrophic Forgetting in Neural Networks},
  author = {Kirkpatrick, James and Pascanu, Razvan and Rabinowitz, Neil and Veness, Joel and Desjardins, Guillaume and Rusu, Andrei A. and Milan, Kieran and Quan, John and Ramalho, Tiago and {Grabska-Barwinska}, Agnieszka and Hassabis, Demis and Clopath, Claudia and Kumaran, Dharshan and Hadsell, Raia},
  year = 2017,
  journal = {Proceedings of the National Academy of Sciences},
  volume = {114},
  number = {13},
  pages = {3521--3526}
}

@inproceedings{kumar_maintaining_2025,
  title = {Maintaining Plasticity in Continual Learning via Regenerative Regularization},
  booktitle = {Proceedings of the 3rd Conference on Lifelong Learning Agents},
  author = {Kumar, Saurabh and Marklund, Henrik and Roy, Benjamin Van},
  editor = {Lomonaco, Vincenzo and Melacci, Stefano and Tuytelaars, Tinne and Chandar, Sarath and Pascanu, Razvan},
  year = 2025,
  month = jul,
  series = {Proceedings of Machine Learning Research},
  volume = {274},
  pages = {410--430},
  publisher = {PMLR}
}

@inproceedings{levine_learning_2017,
  title = {Learning Hand-Eye Coordination for Robotic Grasping with Deep Learning and Large-Scale Data Collection},
  booktitle = {International {{Symposium}} on {{Experimental Robotics}}},
  author = {Levine, Sergey and Sampedro, Peter Pastor and Krizhevsky, Alex and Ibarz, Julian and Quillen, Deirdre},
  year = 2017
}

@inproceedings{liu_definition_2023,
  title = {A {{Definition}} of {{Non-Stationary Bandits}}},
  booktitle = {{{arXiv}}:2302.12202},
  author = {Liu, Yueyang and Kuang, Xu and Roy, Benjamin Van},
  year = 2023,
  month = jul,
  eprint = {2302.12202},
  primaryclass = {cs},
  publisher = {arXiv},
  urldate = {2025-10-17},
  archiveprefix = {arXiv},
  langid = {english}
}

@article{liu_learning_2025,
  title = {Learning Latent and Changing Dynamics in Real Non-Stationary Environments},
  author = {Liu, Zihe and Lu, Jie and Xuan, Junyu and Zhang, Guangquan},
  year = 2025,
  journal = {IEEE Transactions on Knowledge and Data Engineering},
  volume = {37},
  number = {4},
  pages = {1930--1942}
}

@inproceedings{marlos_c_machado_domain-independent_2015,
  title = {Domain-{{Independent Optimistic Initialization}} for {{Reinforcement Learning}}},
  booktitle = {Learning for {{General Competency}} in {{Video Games}}: {{Papers}} from the 2015 {{AAAI Workshop}}},
  author = {Machado, Marlos C. and Srinivasan, Sriram and Bowling, Michael},
  year = 2015
}

@article{mnih_human-level_2015,
  title = {Human-Level Control through Deep Reinforcement Learning},
  author = {Mnih, Volodymyr and Kavukcuoglu, Koray and Silver, David and Rusu, Andrei A. and Veness, Joel and Bellemare, Marc G. and Graves, Alex and Riedmiller, Martin and Fidjeland, Andreas K. and Ostrovski, Georg and Petersen, Stig and Beattie, Charles and Sadik, Amir and Antonoglou, Ioannis and King, Helen and Kumaran, Dharshan and Wierstra, Daan and Legg, Shane and Hassabis, Demis},
  year = 2015,
  month = feb,
  journal = {Nature},
  volume = {518},
  number = {7540},
  pages = {529--533},
  issn = {1476-4687}
}

@inproceedings{nikishin_primacy_2022,
  title = {The Primacy Bias in Deep Reinforcement Learning},
  booktitle = {Proceedings of the 39th International Conference on Machine Learning},
  author = {Nikishin, Evgenii and Schwarzer, Max and D'Oro, Pierluca and Bacon, Pierre-Luc and Courville, Aaron},
  editor = {Chaudhuri, Kamalika and Jegelka, Stefanie and Song, Le and Szepesvari, Csaba and Niu, Gang and Sabato, Sivan},
  year = 2022,
  month = jul,
  series = {Proceedings of Machine Learning Research},
  volume = {162},
  pages = {16828--16847},
  publisher = {PMLR}
}

@inproceedings{osband_behaviour_2020,
  title = {Behaviour Suite for Reinforcement Learning},
  booktitle = {International Conference on Learning Representations},
  author = {Osband, Ian and Doron, Yotam and Hessel, Matteo and Aslanides, John and Sezener, Eren and Saraiva, Andre and McKinney, Katrina and Lattimore, Tor and Szepesv{\'a}ri, Csaba and Singh, Satinder and Van Roy, Benjamin and Sutton, Richard and Silver, David and {van Hasselt}, Hado},
  year = 2020
}

@article{osband_deep_2019,
  title = {Deep Exploration via Randomized Value Functions},
  author = {Osband, Ian and Roy, Benjamin Van and Russo, Daniel J. and Wen, Zheng},
  year = 2019,
  journal = {Journal of Machine Learning Research},
  volume = {20},
  number = {124},
  pages = {1--62}
}

@misc{raffin_rl_2020,
  title = {{{RL Baselines3 Zoo}}},
  author = {Raffin, Antonin},
  year = 2020,
  publisher = {GitHub}
}

@inproceedings{raffin_smooth_2022,
  title = {Smooth Exploration for Robotic Reinforcement Learning},
  booktitle = {Proceedings of the 5th Conference on Robot Learning},
  author = {Raffin, Antonin and Kober, Jens and Stulp, Freek},
  editor = {Faust, Aleksandra and Hsu, David and Neumann, Gerhard},
  year = 2022,
  month = nov,
  series = {Proceedings of Machine Learning Research},
  volume = {164},
  pages = {1634--1644},
  publisher = {PMLR}
}

@article{raffin_stable-baselines3_2021,
  title = {Stable-Baselines3: {{Reliable}} Reinforcement Learning Implementations},
  author = {Raffin, Antonin and Hill, Ashley and Gleave, Adam and Kanervisto, Anssi and Ernestus, Maximilian and Dormann, Noah},
  year = 2021,
  journal = {Journal of Machine Learning Research},
  volume = {22},
  number = {268},
  pages = {1--8}
}

@article{richards_persistence_2017,
  title = {The {{Persistence}} and {{Transience}} of {{Memory}}},
  author = {Richards, Blake A. and Frankland, Paul W.},
  year = 2017,
  month = jun,
  journal = {Neuron},
  volume = {94},
  number = {6},
  pages = {1071--1084},
  issn = {1097-4199},
  langid = {english},
  pmid = {28641107}
}

@inproceedings{rolnick_experience_2019,
  title = {Experience Replay for Continual Learning},
  booktitle = {Advances in Neural Information Processing Systems},
  author = {Rolnick, David and Ahuja, Arun and Schwarz, Jonathan and Lillicrap, Timothy and Wayne, Gregory},
  editor = {Wallach, H. and Larochelle, H. and Beygelzimer, A. and {dAlch{\'e}-Buc}, F. and Fox, E. and Garnett, R.},
  year = 2019,
  volume = {32},
  publisher = {Curran Associates, Inc.}
}

@article{ryan_forgetting_2022,
  title = {Forgetting as a Form of Adaptive Engram Cell Plasticity},
  author = {Ryan, Tom{\'a}s J. and Frankland, Paul W.},
  year = 2022,
  month = mar,
  journal = {Nature Reviews Neuroscience},
  volume = {23},
  number = {3},
  pages = {173--186},
  publisher = {Nature Publishing Group},
  issn = {1471-0048},
  urldate = {2026-09-01},
  copyright = {2022 Springer Nature Limited},
  langid = {english}
}

@misc{schulman_proximal_2017,
  title = {Proximal {{Policy Optimization Algorithms}}},
  author = {Schulman, John and Wolski, Filip and Dhariwal, Prafulla and Radford, Alec and Klimov, Oleg},
  year = 2017,
  month = aug,
  number = {arXiv:1707.06347},
  eprint = {1707.06347},
  primaryclass = {cs},
  publisher = {arXiv},
  urldate = {2026-04-25},
  archiveprefix = {arXiv}
}

@article{silver_welcome_2025,
  title = {Welcome to the {{Era}} of {{Experience}}},
  author = {Silver, David and Sutton, Richard S},
  year = 2025,
  langid = {english}
}

@article{strehl_analysis_2008,
  title = {An Analysis of Model-Based Interval Estimation for Markov Decision Processes},
  author = {Strehl, Alexander L. and Littman, Michael L.},
  year = 2008,
  journal = {Journal of Computer and System Sciences},
  volume = {74},
  number = {8},
  pages = {1309--1331},
  issn = {0022-0000}
}

@article{sutton_dyna_1991,
  title = {Dyna, an Integrated Architecture for Learning, Planning, and Reacting},
  author = {Sutton, Richard S.},
  year = 1991,
  month = jul,
  journal = {SIGART Bull.},
  volume = {2},
  number = {4},
  pages = {160--163},
  publisher = {Association for Computing Machinery},
  address = {New York, NY, USA},
  issn = {0163-5719},
  issue_date = {Aug. 1991}
}

@book{sutton_reinforcement_2018,
  title = {Reinforcement Learning: {{An}} Introduction},
  author = {Sutton, Richard S. and Barto, Andrew G.},
  year = 2018,
  edition = {2},
  publisher = {The MIT Press}
}

@inproceedings{sutton_role_2007,
  title = {On the Role of Tracking in Stationary Environments},
  booktitle = {Proceedings of the 24th International Conference on Machine Learning},
  author = {Sutton, Richard S. and Koop, Anna and Silver, David},
  year = 2007,
  series = {Icml '07},
  pages = {871--878},
  publisher = {Association for Computing Machinery},
  address = {New York, NY, USA},
  isbn = {978-1-59593-793-3}
}

@inproceedings{towers_gymnasium_2024,
  title = {Gymnasium: A Standard Interface for Reinforcement Learning Environments},
  booktitle = {Advances in Neural Information Processing Systems},
  author = {Towers, Mark and Kwiatkowski, Ariel and Balis, John and De Cola, Gianluca and Deleu, Tristan and Goul{\~a}o, Manuel and Andreas, Kallinteris and Krimmel, Markus and KG, Arjun and {Perez-Vicente}, Rodrigo and Terry, J and Pierr{\'e}, Andrea and Schulhoff, Sander and Tai, Jun Jet and Tan, Hannah and Younis, Omar G.},
  editor = {Belgrave, D. and Zhang, C. and Lin, H. and Pascanu, R. and Koniusz, P. and Ghassemi, M. and Chen, N.},
  year = 2025,
  volume = {38, Main Conference},
  publisher = {Curran Associates, Inc.}
}

@article{watkins_technical_1992,
  title = {Technical {{Note}}: {{Q-Learning}}},
  author = {Watkins, Christopher J.C.H. and Dayan, Peter},
  year = 1992,
  month = may,
  journal = {Machine Learning},
  volume = {8},
  number = {3},
  pages = {279--292},
  issn = {1573-0565}
}

@inproceedings{weaver_optimal_2001,
  title = {The Optimal Reward Baseline for Gradient-Based Reinforcement Learning},
  booktitle = {Proceedings of the Seventeenth Conference on Uncertainty in Artificial Intelligence},
  author = {Weaver, Lex and Tao, Nigel},
  year = 2001,
  series = {{{UAI}}'01},
  pages = {538--545},
  publisher = {Morgan Kaufmann Publishers Inc.},
  address = {Seattle, Washington},
  isbn = {1-55860-800-1}
}

@article{williams_simple_1992,
  title = {Simple Statistical Gradient-Following Algorithms for Connectionist Reinforcement Learning},
  author = {Williams, Ronald J.},
  year = 1992,
  month = may,
  journal = {Machine Learning},
  volume = {8},
  number = {3},
  pages = {229--256},
  issn = {1573-0565}
}

@inproceedings{xie_deep_2021,
  title = {Deep Reinforcement Learning amidst Continual Structured Non-Stationarity},
  booktitle = {Proceedings of the 38th International Conference on Machine Learning},
  author = {Xie, Annie and Harrison, James and Finn, Chelsea},
  editor = {Meila, Marina and Zhang, Tong},
  year = 2021,
  month = jul,
  series = {Proceedings of Machine Learning Research},
  volume = {139},
  pages = {11393--11403},
  publisher = {PMLR}
}

@inproceedings{yershova_deterministic_2004,
  title = {Deterministic Sampling Methods for Spheres and {{SO}}(3)},
  booktitle = {{{IEEE}} International Conference on Robotics and Automation, 2004. {{Proceedings}}. {{ICRA}} '04. 2004},
  author = {Yershova, A. and LaValle, S.M.},
  year = 2004,
  volume = {4},
  pages = {3974-3980 Vol.4}
}

@inproceedings{zhao_sim--real_2020,
  title = {Sim-to-Real Transfer in Deep Reinforcement Learning for Robotics: A Survey},
  booktitle = {2020 {{IEEE}} Symposium Series on Computational Intelligence ({{SSCI}})},
  author = {Zhao, Wenshuai and Queralta, Jorge Pe{\~n}a and Westerlund, Tomi},
  year = 2020,
  pages = {737--744}
}
	\bibliographystyle{iclr2027_conference}
	
	\appendix
	\section{Proofs}\label{sec_app:proofs}
	\subsection{Proof for \Cref{th:main_th}}
	\begin{proof}
		Consider that only the SsVD update \Cref{eq:SsVD} is applied for $\tau$ consecutive steps starting from the value $Q_{t-\tau}$, where we from now on omit the state-action pair $(s,a)$ for readability. 
		Note that $t_0 = t-\tau$ holds.
		Unrolling the recursion in \Cref{eq:SsVD}, the value of $Q_t$ is given by:
		\begin{align}
			Q_t &= (1-\eta)^\tau Q_{t-\tau} + \eta b \sum_{k=0}^{\tau-1} (1-\eta)^k \\
			&= (1-\eta)^\tau Q_{t-\tau} + \eta b\frac{1-(1-\eta)^\tau}{1-(1-\eta)}\\
			&= (1-\eta)^\tau Q_{t-\tau} + (1-(1-\eta)^\tau)b \label{eq:rewritten}
		\end{align}
		where we used that $\eta\in(0,1]$ and the geometric sum formula.
		For $\eta = 0$, \Cref{eq:rewritten} holds trivially since then $Q_t = Q_{t-\tau}$.
		Further, note that we can denote $Q_t^*$ as a convex combination:
		\begin{equation}\label{eq:convcomb}
			Q_t^* = (1-\eta)^\tau Q_t^* + (1-(1-\eta)^\tau)Q_t^*.
		\end{equation}
		The goal is now to get an expression for the difference of the action-value function to the optimal action-value function at timestep $t$ if it was not updated for $\tau$ steps by the classic Q-learning rule.
		Denote the estimation error at time $t-\tau$ as $\epsilon_{t-\tau} = Q_{t-\tau} - Q^*_{t-\tau}$.
		Then the difference is given by:
		\begin{align}\label{eq:Qrew}
			Q_t - Q^*_t &= (1-\eta)^\tau Q_{t-\tau} + (1-(1-\eta)^\tau)b - Q^*_t \quad\text{(using \Cref{eq:rewritten}})\nonumber\\
			&= (1-\eta)^\tau (Q^*_{t-\tau} + \epsilon_{t-\tau}) + (1-(1-\eta)^\tau)b - Q^*_t\nonumber \\
			&= (1-\eta)^\tau (Q^*_{t-\tau} + \epsilon_{t-\tau}) + (1-(1-\eta)^\tau)b \nonumber\\ &\phantom{=, }- (1-\eta)^\tau Q_t^* - (1-(1-\eta)^\tau)Q_t^* \quad\text{(using \Cref{eq:convcomb}})\nonumber\\ 
			&= (1-\eta)^\tau (Q^*_{t-\tau} -  Q_t^* + \epsilon_{t-\tau}) + (1-(1-\eta)^\tau)(b- Q_t^*) 
		\end{align}
		which was to be shown. 
	\end{proof}
	
	\subsection{Proof for \Cref{corr:norm}}
	\begin{proof}
		As before, we omit the state-action pair $(s,a)$ for readability. Taking absolute
		values in \Cref{eq:Qrew} and applying the triangle inequality, using that
		$(1-\eta)^\tau \ge 0$ and $1-(1-\eta)^\tau \ge 0$, yields
		\[
		|Q_t - Q^*_t| \le (1-\eta)^\tau \big(|Q^*_{t-\tau} - Q^*_t| + |\epsilon_{t-\tau}|\big)
		+ (1-(1-\eta)^\tau)\,|b - Q^*_t|.
		\]
		By \Cref{eq:assumption} applied to each term using again the triangle inequality this yields 
		\[
		|Q_t - Q^*_t| \le (1-(1-\eta)^\tau)\,|Q^*_t - b| + (1-\eta)^\tau\big(|\epsilon_{t-\tau}| + \Delta\tau\big),
		\]
		which was to be shown.
	\end{proof}
	
	\subsection{Proof for \Cref{corr:optimal_eta}}
	\begin{proof}
		The bound in \Cref{corr:norm} can be written as
		\[
		(1-(1-\eta)^\tau) D + (1-\eta)^\tau S = D + (1-\eta)^\tau (S - D).
		\]
		For $\tau \ge 1$, $(1-\eta)^\tau$ is strictly decreasing in $\eta$ on $[0,1]$,
		with value $1$ at $\eta=0$ and $0$ at $\eta=1$. Hence, if $S > D$, the bound is
		minimized for $\eta = 1$, and if $S < D$, it is minimized for $\eta = 0$.
		If $S = D$, the bound is independent of $\eta$.
	\end{proof}
	\subsection{Proof for \Cref{eq:op_init_val}}
	\begin{proof}
		As before, we omit the state-action pair $(s,a)$ for readability. Rearranging
		\Cref{eq:rewritten} gives
		\[
		Q_t - b = (1-\eta)^\tau \big(Q_{t-\tau} - b\big) = (1-\eta)^\tau \big(Q_{t_0} - b\big).
		\]
		Since no Q-learning update occurs after $t_0$, we have $\tau = t - t_0 \to \infty$
		as $t \to \infty$. As $0 < 1-\eta < 1$ and $Q_{t_0}$ is finite, the right-hand
		side converges to $0$, which was to be shown.
		
		The same holds if the SsVD update is only applied to $(s,a)$ independently with
		some probability $p > 0$ in each timestep, in which case the limit holds almost surely.
	\end{proof}
	
	\section{Additional Experimental Details} \label{sec:app_add_det}
	
	\begin{figure}[b!]
		\centering
		\includegraphics[width=0.95\columnwidth]{
			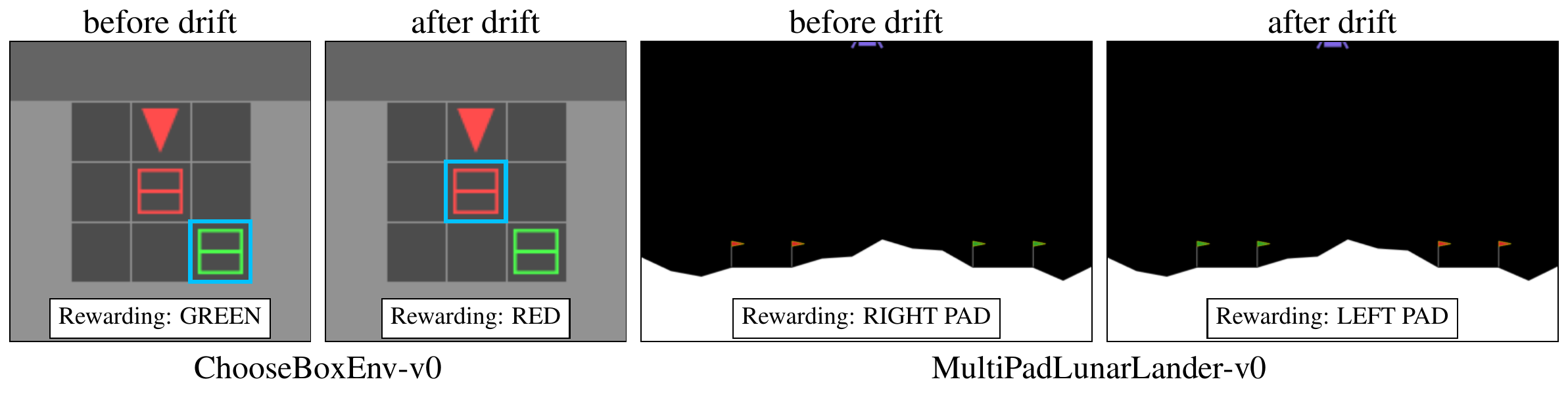} 
		\caption{Visualization of two custom non-stationary environments. \env{ChooseBox} a minigrid environment with switching boxes (left) and \env{MultiPadLunarLander} which extends the classic \env{LunarLander} by an additional landing pad (right). }
		\label{fig:custom_nsrl_envs}
	\end{figure}
	
	\subsection{Custom Non-Stationary Environments}\label{sec:custom_envs}
	We introduce two novel non-stationary environments, both are shown in \Cref{fig:custom_nsrl_envs} together with an example of how these environment drift. 
	
	\textbf{MultiPadLunarLander} extends the established \env{LunarLander} environment \cite{towers_gymnasium_2024} with a larger area and a second landing pad. The target pad switches with a given probability. The agent receives a positive bonus for landing safely on the target pad, a penalty for crashing, and neither if it hovers until the episode terminates. Since we modify only the position of the target pad, the default shaping rewards of the original environment are retained on top of these terminal rewards.
	Similar to the original, the observation space remains $8$ dimensional and the action space can either be discrete ($4$ actions) or continuous ($2$ dimensions for the booster strengths of the side and main engines) -- we benchmark both variants.

	\textbf{ChooseBox} is a custom MiniGrid environment \cite{chevalier-boisvert_minigrid_2023} in which the agent must pick up one of two boxes. At predefined timesteps, the target box switches, so the agent has to learn new behavior. Picking up a box yields a reward of $a \cdot \left(10 - 9\,\frac{t}{T}\right)$, where $t$ is the current step, $T$ is the maximum episode length, and $a = 1$ for the correct box and $a = -1$ for the incorrect one. The reward therefore decays linearly from $10$ to $1$ over the course of an episode, which favors fast solutions. Truncated episodes yield a reward of zero.
	
	Neither environment signals the change to the agent. In \env{ChooseBox}, the boxes swap roles without changing color, so the image-like observations (see \Cref{sec_app:sampling}) look exactly the same before and after a switch. In \env{MultiPadLunarLander}, the target pad does change color, but color is not part of the observation. However, because the shaping rewards depend on the target position, the agent is informed of a switch fairly quickly through the reward signal.

	\subsection{Forms of Non-stationarity}\label{sec_app:non_st_ex}
	\begin{figure}[h]
		\centering
		\includegraphics[width=1\columnwidth]{
			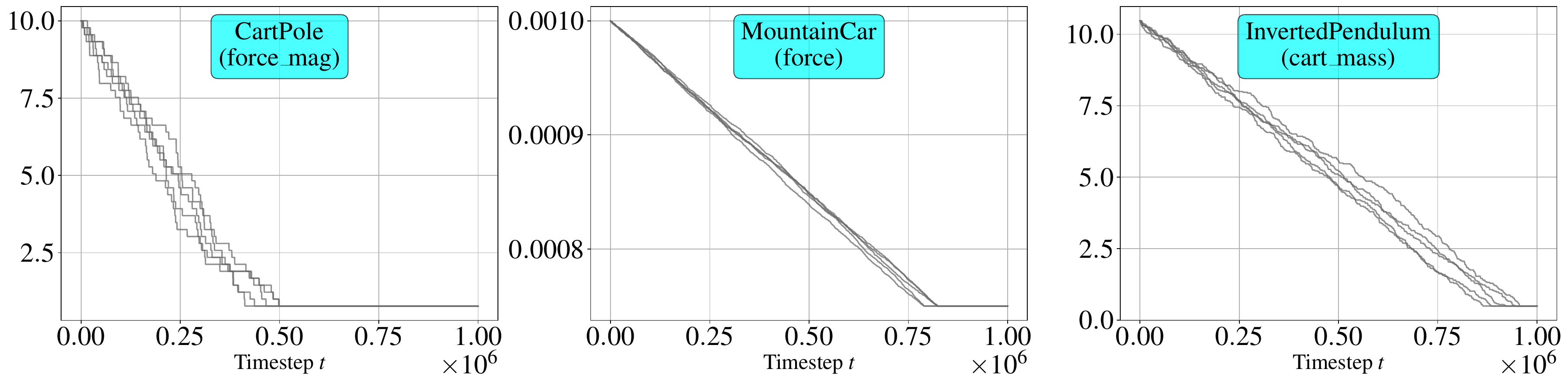} 
		\caption{Trajectories with different seeds of the induced non-stationarity on each environment.}
		\label{fig:Drift Overview}
	\end{figure}
	
	For the newly introduced custom environments see \Cref{sec:custom_envs}.
	For the other environments, the implementation of changing parameters is done with NS-Gym \citep{keplinger_ns-gym_2025-1} which provides tools to \emph{schedule} drifts (when and how often these occur) and to define the performed \emph{update} when a scheduler fires.
	In all cases we use induced changes that retain across episodes.
	We utilize a probabilistic scheduler and incremental updates which were customized with upper/lower bounds.
	In other words, at every timestep with a given probability an incremental modification of the dynamics (see \Cref{tab:drift_environments} for which exact parameters are modified) is applied where due to the probabilistic nature we include bounds on the maximum allowed change (e.g. decreasing gravity should not become zero).
	A handful of different drifting trajectories per environment are shown in \Cref{fig:Drift Overview} with the full details on the parametrizations of these listed in \Cref{tab:scheduler-updater}.\\
	
	For \env{MultiPadLunarLander} the chosen drift probability yields an expected number of 8 flips across the duration of $10^6$ timesteps: note that this exact number is different for each run, with the code on GitHub (see \Cref{sec:exps}) we additionally provide the original result files as ``.json'' format which records the exact timesteps.

	\begin{table}[t]
		\centering
		\resizebox{\textwidth}{!}{
			\begin{tabular}{l|l|l|c|c}
				\hline
				\textbf{Environment} & \textbf{Scheduler} & \textbf{Update Function} & \textbf{Parameter} & \textbf{Initial Value} \\
				\hline
				CartPole & Random($p=0.00009$) & BoundedDecrement($k$ = 0.225, $\text{lo} = 0.9$) & \texttt{force\_mag} & $10.0$ \\
				MountainCar & Random($p=0.0025$) & BoundedDecrement(
				$k= 0.000000125$, $\text{lo}=0.00075$) & \texttt{force} & $0.001$ \\
				\midrule
				InvertedPendulum & Random($p=0.0004$) & BoundedDecrement($k=0.0275$, $\text{lo}=0.5$) & \texttt{cart\_mass} & $\approx10.47$  \\
				\midrule
				MultiPadLunarLander (discr./cont.) & Random($p=0.000008$) & (shaping) reward flip & \texttt{reward} & --  \\
				ChooseBox & Fixed $[5,10,15] \cdot 10^5$ & reward flip & \texttt{reward} & --  \\
				\hline
			\end{tabular}
		}
		\caption{Scheduler configuration and update functions. ``lo'' denotes lower bound. The affected environment parameter and its initial parameter values (at $t=0$) are listed.}
		\label{tab:scheduler-updater}
	\end{table}
	
	\subsection{Special Exploration Environments}\label{sec:exploration_envs}
	
	\begin{figure}[h!]
		\centering
		\definecolor{envRight}{RGB}{0,114,178}
		\definecolor{envLeft}{RGB}{213,94,0}
		\definecolor{envGold}{RGB}{230,159,0}
		\begin{tikzpicture}[
			>=stealth,
			rsstate/.style={circle, draw=black!80, thick, fill=white, minimum size=9mm, inner sep=0pt},
			rsR/.style={->, semithick, draw=envRight},
			rsL/.style={->, semithick, dashed, draw=envLeft},
			rsp/.style={font=\scriptsize, text=envRight, inner sep=1.5pt},
			]
			% states
			\foreach \i in {0,...,5}
			\node[rsstate] (s\i) at ({1.88*\i},0) {$s_{\i}$};
			
			% RIGHT: moves right (above)
			\draw[rsR] (s0) to[bend left=45] node[rsp, above] {0.40} (s1);
			\foreach \i/\j in {1/2,2/3,3/4,4/5}
			\draw[rsR] (s\i) to[bend left=45] node[rsp, above] {0.35} (s\j);
			% RIGHT: stays (self-loops)
			\foreach \i in {0,...,5}
			\draw[rsR] (s\i.100) .. controls +(110:1.05) and +(70:1.05) .. node[rsp, above] {0.60} (s\i.80);
			% RIGHT: pushed back (below)
			\foreach \i/\j in {1/0,2/1,3/2,4/3}
			\draw[rsR] (s\i) to[bend left=45] node[rsp, below] {0.05} (s\j);
			\draw[rsR] (s5) to[bend left=45] node[rsp, below] {0.40} (s4);
			% LEFT: deterministic
			\foreach \i/\j in {1/0,2/1,3/2,4/3,5/4}
			\draw[rsL] (s\i) -- (s\j);
			\draw[rsL] (s0.190) .. controls +(200:1.05) and +(160:1.05) .. (s0.170);
			
			% rewards
			\node[font=\small, text=envLeft, anchor=east] at (-1.25,0) {$r=5$};
			\node[font=\small, text=envRight, anchor=west] at ([xshift=1.2mm]s5.east) {$r=10^{4}$};
			
			% start marker
			\draw[<-, thick, black!80] (s0.south) -- ++(0,-0.75) node[below, font=\small, text=black] {start};
			
			% legend
			\begin{scope}[shift={(4.9,-1.35)}]
				\draw[rsR] (-2.3,0) -- (-1.5,0) node[right, font=\small, text=black] {\textsc{right}};
				\draw[rsL] (0.4,0) -- (1.2,0) node[right, font=\small, text=black] {\textsc{left} (prob.\ 1)};
			\end{scope}
		\end{tikzpicture}
		\caption{RiverSwim with $S=6$ states. Solid blue arrows show transitions under \textsc{right} with their probabilities; dashed orange arrows show \textsc{left}, which is deterministic. Choosing \textsc{left} in $s_0$ pays $5$ and choosing \textsc{right} in $s_5$ pays $10^{4}$ (regardless of the resulting transition); zero otherwise.}
		\label{fig:riverswim}
	\end{figure}
	
	\paragraph{RiverSwim}
	 \citep{strehl_analysis_2008} is a chain of $S=6$ states $s_0,\dots,s_5$ in a river that flows to the left (Figure~\ref{fig:riverswim}). Swimming \textsc{left}, with the current, always succeeds, whereas swimming \textsc{right}, against the current, advances the swimmer only with probability $0.35$ ($0.40$ in $s_0$) and otherwise leaves it in place or pushes it back. As in the original formulation, the agent receives a reward of $5$ for choosing \textsc{left} in $s_0$ and $10^{4}$ for choosing \textsc{right} in $s_5$; all other rewards are zero. Episodes start in $s_0$ and are truncated after $H=200$ steps,  the agent observes a one-hot encoding of its state.
	
	\env{RiverSwim} tests deep exploration under stochastic transitions. The small reward in $s_0$ is available immediately and yields a return of $1000$ per episode, whereas the large reward requires swimming against the current for many consecutive steps without any intermediate feedback. An agent that dithers around the locally optimal ``always \textsc{left}'' policy is carried back by the current and almost never reaches $s_5$.

	\paragraph{DeepSea} \cite{osband_deep_2019}
	 is an $N\times N$ grid with different values for $N$; we follow the \texttt{bsuite} implementation \citep{osband_behaviour_2020} (Figure~\ref{fig:deepsea}). The agent starts in the top-left cell and in every step descends one row while moving one column to the right or to the left, so each episode terminates after exactly $N$ steps. Which of the two actions moves right is drawn at random for every cell from a fixed seed and stays the same across episodes. Every move to the right costs $0.01/N$, and moving right in the bottom-right cell additionally yields a treasure of $+1$. The agent observes a one-hot encoding of its cell ($N^2$ dimensions).
	
	\env{DeepSea} isolates the need for deep exploration in a deterministic environment. Because the horizontal position can increase by at most one per step, the treasure is obtained by exactly one of the $2^N$ action sequences, moving right in every step, which yields the optimal return of $0.99$. Every other sequence returns $-0.01\,k/N\le 0$, where $k$ is the number of moves to the right, so until the treasure has been found the move cost makes moving left look optimal everywhere. A uniformly random policy finds the treasure with probability $2^{-N}$ per episode, so dithering strategies need on the order of $2^{N}$ episodes, and the per-cell randomization of actions rules out shortcuts such as always choosing the same action.
	
	%  DeepSea 
	\begin{figure}[t!]
		\centering
		\definecolor{envRight}{RGB}{0,114,178}
		\definecolor{envLeft}{RGB}{213,94,0}
		\definecolor{envGold}{RGB}{230,159,0}
		\begin{tikzpicture}[
			x=5.5mm, y=5.5mm,
			>=stealth,
			dsdigit/.style={font=\tiny, text=black!60, inner sep=0pt},
			]
			% unreachable cells (j > i)
			\foreach \i in {0,...,8} {
				\pgfmathtruncatemacro{\jmin}{\i+1}
				\foreach \j in {\jmin,...,9}
				\fill[black!10] (\j,-\i) rectangle ++(1,-1);
			}
			% optimal path and treasure
			\foreach \i in {0,...,8}
			\fill[envRight!20] (\i,-\i) rectangle ++(1,-1);
			\fill[envGold!60] (9,-9) rectangle ++(1,-1);
			% grid
			\foreach \k in {1,...,9} {
				\draw[black!30, very thin] (\k,0) -- (\k,-10);
				\draw[black!30, very thin] (0,-\k) -- (10,-\k);
			}
			\draw[black!75] (0,0) rectangle (10,-10);
			
			% index of the action that moves right, a^->(i,j), for the default mapping
			% (N = 10, mapping_seed = 42); reachable cells only. Delete these lines to hide the digits.
			\foreach \j/\a in {0/0} \node[dsdigit] at (\j+0.5,-0.5) {\a};
			\foreach \j/\a in {0/1,1/1} \node[dsdigit] at (\j+0.5,-1.5) {\a};
			\foreach \j/\a in {0/1,1/0,2/0} \node[dsdigit] at (\j+0.5,-2.5) {\a};
			\foreach \j/\a in {0/0,1/0,2/0,3/1} \node[dsdigit] at (\j+0.5,-3.5) {\a};
			\foreach \j/\a in {0/0,1/1,2/1,3/0,4/0} \node[dsdigit] at (\j+0.5,-4.5) {\a};
			\foreach \j/\a in {0/1,1/0,2/0,3/0,4/0,5/0} \node[dsdigit] at (\j+0.5,-5.5) {\a};
			\foreach \j/\a in {0/1,1/1,2/0,3/1,4/0,5/0,6/1} \node[dsdigit] at (\j+0.5,-6.5) {\a};
			\foreach \j/\a in {0/1,1/0,2/0,3/0,4/1,5/0,6/0,7/0} \node[dsdigit] at (\j+0.5,-7.5) {\a};
			\foreach \j/\a in {0/1,1/0,2/0,3/1,4/1,5/1,6/0,7/0,8/1} \node[dsdigit] at (\j+0.5,-8.5) {\a};
			\foreach \j/\a in {0/0,1/1,2/1,3/0,4/1,5/0,6/0,7/1,8/1,9/0} \node[dsdigit] at (\j+0.5,-9.5) {\a};
			
			% optimal trajectory
			\foreach \i in {0,...,8}
			\draw[->, thick, envRight] (\i+0.72,-\i-0.72) -- (\i+1.28,-\i-1.28);
			\draw[->, thick, envRight] (9.72,-9.72) -- (10.55,-10.55)
			node[anchor=north west, inner sep=1pt, font=\small, text=black] {$+1$};
			
			% axes and start
			\draw[->, black!60] (0,0.7) -- (10,0.7) node[midway, above, font=\small, text=black] {column $j$};
			\draw[->, black!60] (-0.7,0) -- (-0.7,-10);
			\node[rotate=90, font=\small] at (-1.45,-5) {row $i$ (= time step)};
			\draw[->, thick, black!80] (-1.5,1.5) node[above left, inner sep=1pt, font=\small] {start} -- (0.08,-0.08);
			
			% inset: one step of the dynamics
			\begin{scope}[shift={(7.4cm,-2.0cm)}, x=1cm, y=1cm,
				mini/.style={draw=black!70, minimum width=15mm, minimum height=7mm, inner sep=2pt, font=\scriptsize}]
				\node[mini] (c)  at (1.45,0)    {$(i,j)$};
				\node[mini] (bl) at (-0.1,-1.45)   {$(i{+}1,\,j{-}1)$};
				\node[mini] (br) at (3.0,-1.45) {$(i{+}1,\,j{+}1)$};
				\draw[->, thick, envRight] (c.south) ++(0.2,0) -- (br.north);
				\draw[->, thick, dashed, envLeft] (c.south) ++(-0.2,0) -- (bl.north);
				\node[font=\scriptsize, text=envLeft, align=center, anchor=north] at (bl.south) {move left\\$r=0$};
				\node[font=\scriptsize, text=envRight, align=center, anchor=north] at (br.south) {move right\\$r=-0.01/N$};
				\node[font=\small, anchor=south] at (1.45,0.45) {one step from $(i,j)$};
			\end{scope}
		\end{tikzpicture}
		\caption{DeepSea with $N=10$. The agent starts in the top-left cell and descends one row per step while moving one column left or right. Grey cells are unreachable. The only rewarding trajectory is the diagonal (blue), which moves right in every step and collects the treasure ($+1$) by moving right in the bottom-right cell (gold). Digits show which action index moves right in each cell.}
		\label{fig:deepsea}
	\end{figure}
	
	\subsection{Sampling}\label{sec_app:sampling} 
	
	\paragraph*{ChooseBox} The observations in this environment (see \Cref{sec:custom_envs}) are encoded with the Minigrid default as a symbolic $7\times7$ grid where each cell contains a triple of integers (object\_idx, color\_idx, state) which encode the type, color and state of the observed cell.
	Clearly, naively sampling states is not possible: instead we observe that the amount of possible game states is finite ($4$ agent orientations, $9$ positions for the agent, $8$ for the first box and thus $7$ possible choices for the second box, yielding $9\cdot8\cdot7\cdot4=2016$ different states) and we can use the Minigrid interface to generate one of these possible game states and use these as samples.
	
	\paragraph*{Others} The observation and action spaces of the other environments are constrained and thus we rely on the built-in sampling methods of the defined observation spaces \cite{towers_gymnasium_2024}.
	
	\begin{table}[b]
		\centering
		\small
		\caption{Mean online return per environment, averaged over seeds, with the within-environment rank in parentheses (best in bold). Returns are only comparable within a row.}
		\label{tab_app:mor}
		\resizebox{\textwidth}{!}{
			\begin{tabular}{lrrrrrr}
				\toprule
				Environment & Base (DQN/SAC) & F (ours) & L2ToInit & Limited & PPO & Resetting  \\
				\midrule
				CartPole-v1 & 98.9 (5) & 394.3 (2) & 108.3 (4) & 48.7 (6) & \textbf{475.5 (1)} & 213.8 (3) \\
				ChooseBoxEnv-v0 & 0.7 (2) & \textbf{1.6 (1)} & -0.1 (5) & -0.6 (6) & 0.1 (4) & 0.4 (3) \\
				InvertedPendulum-v5 & 945.0 (2) & \textbf{984.1 (1)} & 901.1 (4) & 390.2 (6) & 860.1 (5) & 911.0 (3) \\
				MountainCar-v0 & -187.3 (4) & -177.3 (3) & -171.9 (2) & -200.0 (6) & \textbf{-147.8 (1)} & -187.3 (5) \\
				MultiPadLunarLander (cont.) & -25.7 (5) & -0.3 (3) & \textbf{1.5 (1)} & -37.0 (6) & -0.2 (2) & -4.1 (4) \\
				MultiPadLunarLander (discr.) & -280.6 (6) & 14.3 (2) & 6.2 (3) & -110.9 (5) & \textbf{24.1 (1)} & 1.8 (4) \\
				\bottomrule
			\end{tabular}
	}\end{table}

	\subsection{Hyperparameters}\label{sec_app:hpo}
	\paragraph*{Non-stationary}(\Cref{sec:emp_ana_nsrl})
	All used hyper-parameters are found in \Cref{tab:hyperparams} -- the base parameters were chosen according to the tuned parameter values found in RL zoo \cite{raffin_rl_2020}.
	The reset interval for \citet{nikishin_primacy_2022}'s approach is chosen heuristically; additionally we use an $\varepsilon$ warmup strategy for DQN: this overrides the default chosen exploration strategy by applying the original schedule after each reset (including an initial reset at $t=0$) so that the final \texttt{exploration\_final\_eps} is reached again after $10^4$ steps.
	The regularization strength for L2ToInit \cite{kumar_maintaining_2025} is determined by running the algorithm for $\lambda\in \{10^{-3},10^{-4},10^{-5}\}$ on each environment and picking the best value based on the result.
	The parameters for our proposed forgetting mechanism SsVD were also chosen heuristically testing a few variations in magnitude for the parameter $\eta$ and selected values for $b$.
	Note that the baseline $b=-50$ for \env{MultiPadLunarLander} can be seen as either optimistic or pessimistic (see \Cref{sec:custom_envs}).
	 
	\begin{figure*}[b!]
		\centering
		\includegraphics[width=1\textwidth]{
			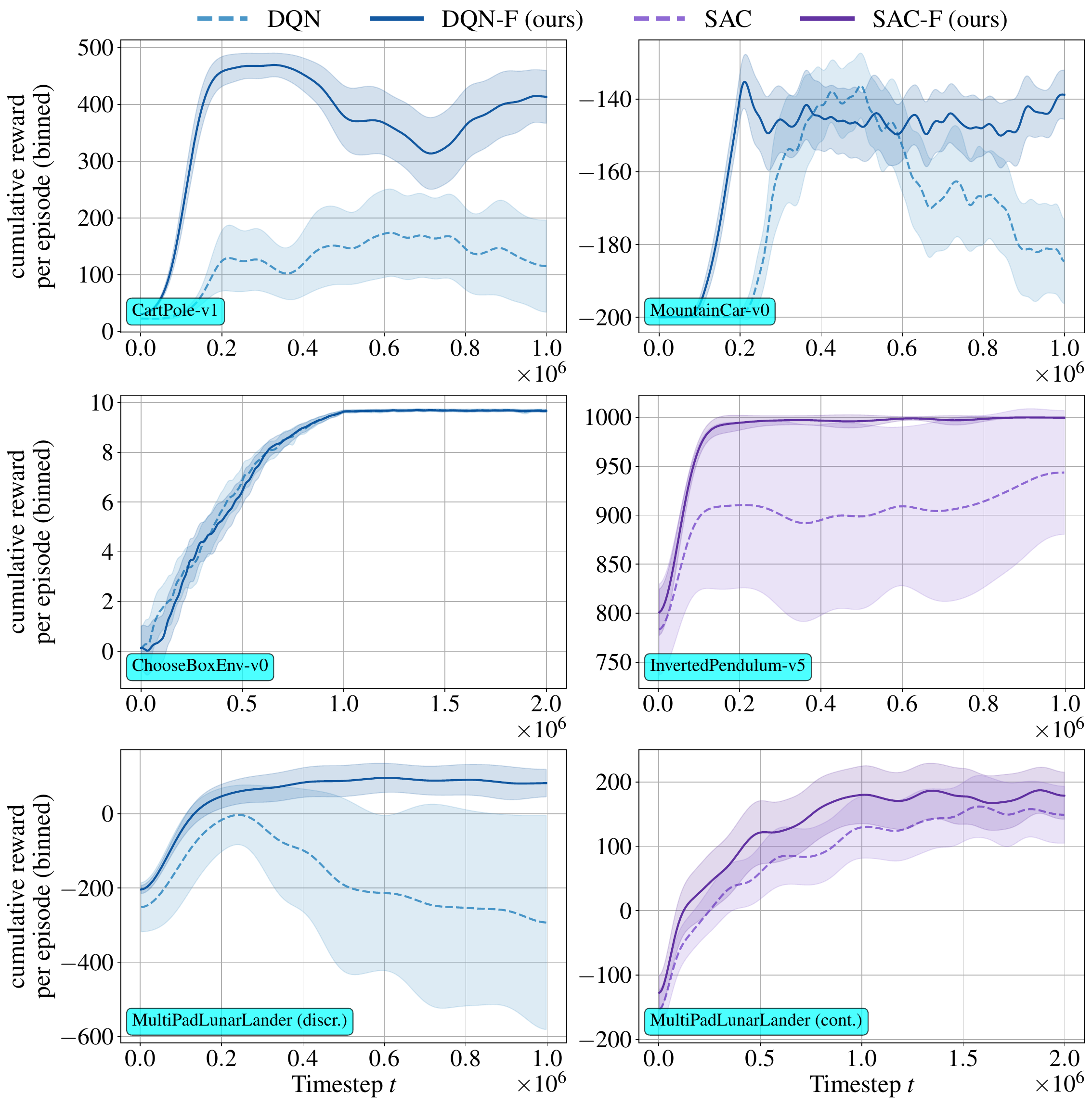}
		\caption{Same experimental setup as \Cref{fig:main_result} but with stationary environments. Only the base algorithm (SAC or DQN) and their counterpart using our proposed SsVD mechanisms are shown.}
		\label{fig_app:stationary_ablation}
	\end{figure*}
	
	\paragraph*{Exploration}(\Cref{sec:hard_exploration})
	We recall the hyperparameters of SsVD (see \Cref{sec:main_method} for details): the forgetting ratio $\xi$, the forgetting learning rate $\eta$, and the baseline value $b$. Some of these follow a linear schedule, denoted $\mathrm{Lin}(a; b; c)$: the value starts at $a$ and decays linearly towards $b$ until a fraction $c$ of the total training time has passed, after which it remains at $b$.
	
	On \env{MountainCarContinuous}, we use $\eta = \mathrm{Lin}(0.15; 0; 0.5)$ and $\xi = \mathrm{Lin}(0.25; 0; 0.5)$, so SsVD becomes inactive after $50\%$ of training. The baseline is fixed and chosen optimistically as $b = 100$.
	On \env{RiverSwim}, we use fixed values $b = 5000$ and $\eta = 0.25$, together with $\xi = \mathrm{Lin}(1; 0; 0.25)$.
	On \env{DeepSea}, we use a fixed $\xi = 0.25$, together with $b = \mathrm{Lin}(100; 1; 0.33)$ and $\eta = \mathrm{Lin}(1; 0.005; 0.66)$.

	\section{Additional Visualizations and Results} \label{sec_app:add_res}
	
	\begin{figure}[b!]
		\includegraphics[width=1\textwidth]{
			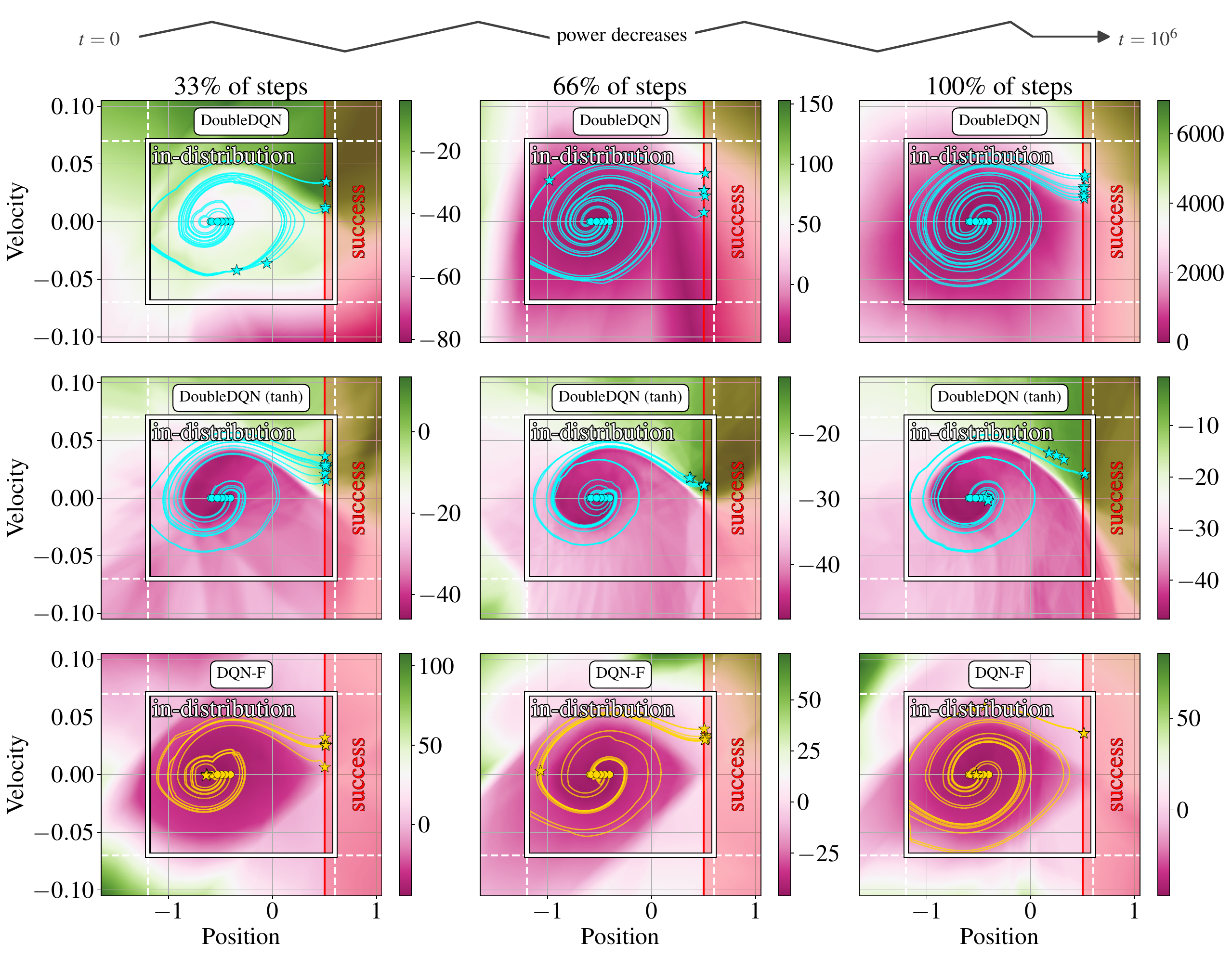}
		\caption{Same experimental setup as \Cref{fig:visabs} but with DoubleDQN instead of DQN and an ablation where the default activation function (ReLU) is replaced by tanh. DoubleDQN is otherwise using the exact same parameters as in \Cref{fig:visabs}.}
		\label{fig_app:doubledqn}
	\end{figure}
	
	\paragraph*{Within-environment Ranking}
	The mean ranking in \Cref{fig:ranking} is based on per-environment rankings of the tested algorithms. We rank within each environment because reward scales differ substantially across environments, which would dominate any direct comparison of raw returns. Within each environment, we rank the algorithms by their mean online return, i.e., the average reward attained over the entire course of training, and then average these ranks across environments. Since the base algorithm differs between environments (DQN or SAC), we treat all limited variants (e.g., Limited DQN and LimitedSAC) as a single ``Limited'' entry, and likewise group the base algorithms into a single entry, enabling a consistent comparison across environments. The per-environment rankings, along with the raw scores for all environments and algorithms, are reported in \Cref{tab_app:mor}.
	
	\begin{table}[b!]
		\centering
		\caption{Hyperparameters used for DQN and SAC. Hyperparameters are used as reported in RL Zoo \cite{raffin_rl_2020}. ``---'' indicates that the parameter does not apply to the method (e.g. exploration\_fraction in SAC). Parameters for the forgetting mechanisms ($\xi,\eta,b$) are found at the bottom and only apply to the -F (SsVD) variants which otherwise use the same parameters as the base algorithm. Number of runs and random seeds used for evaluation is shown as ``num\_iterations''. ``max\_training\_steps'' indicates maximum for ``Limited'' versions. Lin($a$; $b$; $c$) means that the value is initialized with $a$ and linearly decays towards $b$ until a fraction of the total time $c$ has passed. ``gradient\_steps'' $=-1$ means as many gradient steps as environment steps collected per rollout.% (i.e.\ equal to ``train\_freq'')
		}
		\label{tab:hyperparams}
		\setlength{\tabcolsep}{5pt}
		\renewcommand{\arraystretch}{1.3}
		\resizebox{\textwidth}{!}{
			\begin{tabular}{|c|l|cc|c|cc|c|}
				\toprule
				&
				& \textbf{CartPole-v1}
				& \textbf{MountainCar-v0}
				& \textbf{InvertedPendulum-v5}
				& \textbf{MPLL (discr.)}
				& \textbf{MPLL (cont.)}
				& \textbf{ChooseBox} \\
				\midrule
				& \diagbox{\textbf{Hyperparameter}}{\textbf{Algorithm}} 
				& DQN & DQN & SAC & DQN & SAC & DQN \\
				\midrule
				\multirow{15}{*}{\rotatebox[origin=c]{90}{\textbf{Base}}}
				& \texttt{n\_timesteps}
				& $10^{6}$ & $10^{6}$
				& $10^{6}$ & $10^{6}$
				& $10^{6}$ & $2\cdot10^{6}$ \\
				
				& \texttt{policy}
				& MlpPolicy & MlpPolicy
				& MlpPolicy & MlpPolicy
				& MlpPolicy & MlpPolicy \\
				
				& \texttt{learning\_rate}
				& $2.3\cdot10^{-3}$ & $4\cdot10^{-3}$
				& $3\cdot10^{-4}$ & $6.3\cdot10^{-4}$
				& Lin($7.3\cdot10^{-4}$; $0$; $1$) & $10^{-4}$ \\
				
				& \texttt{batch\_size}
				& 64 & 128
				& 256 & 128
				& 256 & 64 \\ 
				
				& \texttt{buffer\_size}
				& 100\_000 & 10\_000
				& 10\_000 & 50\_000
				& 1\_000\_000 & 1\_000\_000 \\ 
				
				& \texttt{gamma} ($\gamma$)
				& 0.995 & 0.98
				& 0.99 & 0.99
				& 0.99 & 0.95 \\ 
				
				& \texttt{tau}
				& 1 & 1
				& 0.005 & 1
				& 0.01 & 1 \\
				
				& \texttt{train\_freq}
				& 256 & 16
				& 1 & 4
				& 1 & 1 \\ 
				
				& \texttt{gradient\_steps}
				& 128 & 8
				& 1 & $-1$
				& 1 & 1 \\ 
				
				& \texttt{target\_update\_interval}
				& 10 (hard) & 600 (hard)
				& 1 (polyak) & 250 (hard)
				& 1 (polyak) & 20 (hard) \\
				
				& \texttt{exploration\_fraction}
				& 0.16 & 0.2 
				& --- & 0.12
				& --- & 0.5 \\ 
				
				& \texttt{exploration\_final\_eps}
				& 0.05 & 0.07
				& --- & 0.1
				& --- & 0.05 \\
				
				& \texttt{ent\_coef}
				& --- & ---
				& \texttt{auto} & ---
				& \texttt{auto} & --- \\ 
				
				& \texttt{use\_sde}
				& --- & ---
				& False & ---
				& False & --- \\
				
				& \texttt{net\_arch}
				& [256, 256] & [256, 256]
				& [256, 256] & [256, 256]
				& [400, 300] & [64, 64] \\
				
				\midrule
				
				\multirow{3}{*}{\rotatebox[origin=c]{90}{\textbf{SsVD}}}
				& \texttt{$\xi$ (for \_F only)}
				& 0.05 & 0.05
				& 0.25 & 0.05
				& 0.05 & 0.25 \\ 
				
				& \texttt{$\eta$ (for \_F only)}
				& 0.015 & 0.0015
				& 0.0015 & 0.015
				& 0.015 & 0.015 \\ 
				
				& \texttt{$b$ (for \_F only)}
				& 0 & 0
				& 0 & 0
				& $-50$ & $5$ \\
				
				\midrule
				
				\multirow{1}{*}{\rotatebox[origin=c]{90}{\textbf{\scriptsize Res.}}}
				
				& \texttt{Interval $J$}
				& $10^5$ & $10^5$
				& $2\cdot10^5$ & $10^5$
				& $2\cdot10^5$ & $4\cdot10^5$ \\
				
				\midrule
				
				\multirow{1}{*}{\rotatebox[origin=c]{90}{\textbf{\scriptsize L2}}}
				& \texttt{$\lambda$}
				& $10^{-5}$ & $10^{-4}$
				& $10^{-5}$ & $10^{-5}$
				& $10^{-4}$ & $10^{-5}$ \\
				
				\midrule
				
				\multirow{1}{*}{\rotatebox[origin=c]{90}{\textbf{\scriptsize Lim.}}}
				& \texttt{max\_training\_steps}
				& $2\cdot10^5$ & $2\cdot10^5$ 
				& $10^5$ & $2\cdot10^5$
				& $2.5\cdot10^5$ & $3.5\cdot10^5$ \\
				
				\midrule
				\midrule
				& \texttt{num\_iterations}
				& 10 & 10
				& 10 & 10
				& 10 & 10 \\
				
				\bottomrule
			\end{tabular}
		}
	\end{table}
	
	\paragraph*{Ablation: Stationary Environments}
	In \Cref{fig_app:stationary_ablation} we illustrate the same empirical evaluation as done in the main text but now the environments are kept stationary.
	We can see that our proposed mechanism \emph{does not harm performance} even if the environments are kept stationary -- in particular our approach even \emph{improves} learning on all environments except on \env{ChooseBox} where both perform equally.
	How strongly an environment is affected by the induced change -- or at least how this exhibits in the reward function -- varies across the tested environments. 
	On \env{InvertedPendulum}, for instance, removing the drift yields little improvement, since our approach already performs near-optimally when it is affected by drift. 
	The environment is nonetheless clearly affected by the drift, as evidenced by the performance drop of the ``Limited'' variant in \Cref{fig:main_result}. 
	In other environments, such as \env{ChooseBox} and \env{MountainCar}, the stationary setting yields clear improvements. 
	This illustrates the heterogeneity of challenges encountered in non-stationary RL.

	\paragraph*{Extended Qualitative Analysis}

	\Cref{fig_app:doubledqn} extends \Cref{fig:visabs} by DoubleDQN~\citep{hasselt_deep_2016}, and DoubleDQN with $\tanh$ instead of ReLU activations; all other parameters are unchanged. Note that each panel has its own color scale. As every step yields $r_t=-1$, all true values lie below $-1$ \emph{independent of the engine power}, so any positive estimate is an overestimate of the true value.
	
	DoubleDQN reduces the final magnitude up to $\approx 6\cdot 10^{3}$, as well as intermediately after $66\%$ of steps. 
	As online and target network are strongly correlated, the double estimator removes noise-induced maximization bias but not such strong divergence.
	With $\tanh$, the values stay within the admissible range throughout, so the divergence appears to require unbounded hidden features.
	As discussed in the main text DQN-F keeps the ReLU architecture without weight decay, yet its values stay at a few tens and nearly all rollouts reach the goal.

\end{document}